\documentclass{article}

\usepackage{amsmath,amsfonts,bm}

\def\eqref#1{equation~\ref{#1}}
\def\1{\bm{1}}

\DeclareMathAlphabet{\mathsfit}{\encodingdefault}{\sfdefault}{m}{sl}
\SetMathAlphabet{\mathsfit}{bold}{\encodingdefault}{\sfdefault}{bx}{n}

\DeclareMathOperator*{\argmax}{arg\,max}

\usepackage{graphicx}  %Required
\usepackage{hyperref}
\usepackage{url}
\usepackage{bm}
\usepackage{multicol}
\usepackage{subfig}
\usepackage{xspace}
\usepackage{xcolor}
\usepackage[export]{adjustbox}
\usepackage{array}
\usepackage{enumitem}

\usepackage{floatrow}
\newfloatcommand{capbtabbox}{table}[][\FBwidth]
\usepackage{blindtext}

\usepackage{multirow}

\newcommand{\srrn}{\textsc{sr-RNN}\xspace}
\newcommand{\srrns}{\textsc{sr-RNN}s\xspace}

\newcommand{\lstm}{\textsc{LSTM}\xspace}
\newcommand{\lstmp}{\textsc{LSTM-p}\xspace}
\newcommand{\lstms}{\textsc{LSTM}s\xspace}
\newcommand{\gru}{\textsc{GRU}\xspace}
\newcommand{\grus}{\textsc{GRU}s\xspace}

\newcommand{\srlstm}{\textsc{sr-LSTM}\xspace}
\newcommand{\srlstms}{\textsc{sr-LSTM}s\xspace}
\newcommand{\srlstmp}{\textsc{sr-LSTM-p}\xspace}
\newcommand{\srlstmps}{\textsc{sr-LSTM-p}s\xspace}

\newcommand{\srgru}{\textsc{sr-GRU}\xspace}
\newcommand{\srgrus}{\textsc{sr-GRU}s\xspace}

\newcommand{\bfb}{\mathbf{b}}
\newcommand{\bfh}{\mathbf{h}}
\newcommand{\bfc}{\mathbf{c}}
\newcommand{\bfx}{\mathbf{x}}
\newcommand{\bfu}{\mathbf{u}}
\newcommand{\bfs}{\mathbf{s}}
\newcommand{\bfp}{\mathbf{p}}
\newcommand{\bff}{\mathbf{f}}
\newcommand{\bfi}{\mathbf{i}}
\newcommand{\bfo}{\mathbf{o}}

\newcommand{\bfW}{\mathbf{W}}
\newcommand{\bfS}{\mathbf{S}}
\newcommand{\bfR}{\mathbf{R}}

\newcommand{\calQ}{\mathcal{Q}}

\usepackage{amsmath, amssymb, amsthm}
\usepackage[algo2e]{algorithm2e}
\usepackage{algorithmic}

\makeatletter
\newcommand{\algrule}[1][.2pt]{\par\vskip.5\baselineskip\hrule height #1\par\vskip.5\baselineskip}
\makeatother

\newtheorem{theorem}{Theorem}[section]

\definecolor{ashgrey}{rgb}{0.5, 0.5, 0.5}
\definecolor{ceruleanblue}{rgb}{0.0, 0.0, 0.0}

\usepackage{color}

\usepackage{graphicx}
\usepackage[colorinlistoftodos]{todonotes}
\usepackage{algorithmic,eqparbox,array}
\renewcommand\algorithmiccomment[1]{%
  \hfill\#\ \eqparbox{COMMENT}{#1}%
}

\usepackage[title]{appendix}
\usepackage{makecell}

\usepackage{microtype}
\usepackage{graphicx}
\usepackage{booktabs} % for professional tables

\usepackage{hyperref}

\usepackage[accepted]{icml2019}

\icmltitlerunning{State-Regularized Recurrent Neural Networks}

\begin{document}

\twocolumn[
\icmltitle{State-Regularized Recurrent Neural Networks}




\begin{icmlauthorlist}
\icmlauthor{Cheng Wang}{to}
\icmlauthor{Mathias Niepert}{to}
\end{icmlauthorlist}

\icmlaffiliation{to}{NEC Laboratories Europe, Heidelberg, Germany}
%\icmlaffiliation{goo}{Googol ShallowMind, New London, Michigan, USA}

\icmlcorrespondingauthor{Cheng Wang}{cheng.wang@neclab.eu}
%\icmlcorrespondingauthor{Eee Pppp}{ep@eden.co.uk}

\icmlkeywords{Machine Learning, ICML}

\vskip 0.3in
]



\printAffiliationsAndNotice{}  % leave blank if no need to mention equal contribution

\begin{abstract}
Recurrent neural networks are a widely used class of neural architectures.  They have, however, two shortcomings. First, it is difficult to understand what exactly they learn. Second, they tend to work poorly on sequences requiring long-term memorization, despite having this capacity in principle. We aim to address both shortcomings with a class of recurrent networks that use a stochastic state transition mechanism between cell applications. This mechanism, which we term state-regularization, makes RNNs transition between a finite set of learnable states. We evaluate state-regularized RNNs on (1) regular languages for the purpose of automata extraction; (2) nonregular languages such as balanced parentheses, palindromes, and the copy task where external memory is required; and (3) real-word sequence learning tasks for sentiment analysis, visual object recognition, and language modeling. We show that state-regularization (a) simplifies the extraction of finite state automata modeling an RNN's state transition dynamics; (b) forces RNNs to operate more like automata with external memory and less like finite state machines; (c) makes RNNs have better interpretability and explainability.
\end{abstract}

\section{Introduction}

Recurrent neural networks (RNNs) have found their way into numerous applications. Still, RNNs have two shortcomings. 
First, it is difficult to understand what concretely RNNs learn. Some applications require a close inspection of learned models before deployment and RNNs are more difficult to interpret than rule-based systems. There are a number of approaches for extracting finite state automata (DFAs) from trained RNNs~\citep{giles1991second,Wang:2007:nc,pmlr-v80-weiss18a} as a means to analyze their behavior. These methods apply extraction algorithms after training and it remains challenging to determine whether the extracted DFA faithfully models the RNN's state transition behavior. Most extraction methods are rather complex, depend crucially on hyperparameter choices, and tend to be computationally costly. 
Second, RNNs tend to work poorly on input sequences requiring long-term memorization, despite having this ability in principle. 
Indeed, there is a growing body of work providing evidence, both empirically~\citep{Daniluk:2017,bai:2018,trinh2018learning} and theoretically~\citep{arjovsky2016unitary,zilly2017recurrent,miller:2018}, that recurrent networks offer no benefit on longer sequences, at least under certain conditions. Intuitively, RNNs tend to operate more like DFAs with a large number of states, attempting to memorize all the information about the input sequence solely with their hidden states, and less like automata with external memory. 

\begin{figure*}%
\centering
\includegraphics[width=0.78\textwidth]{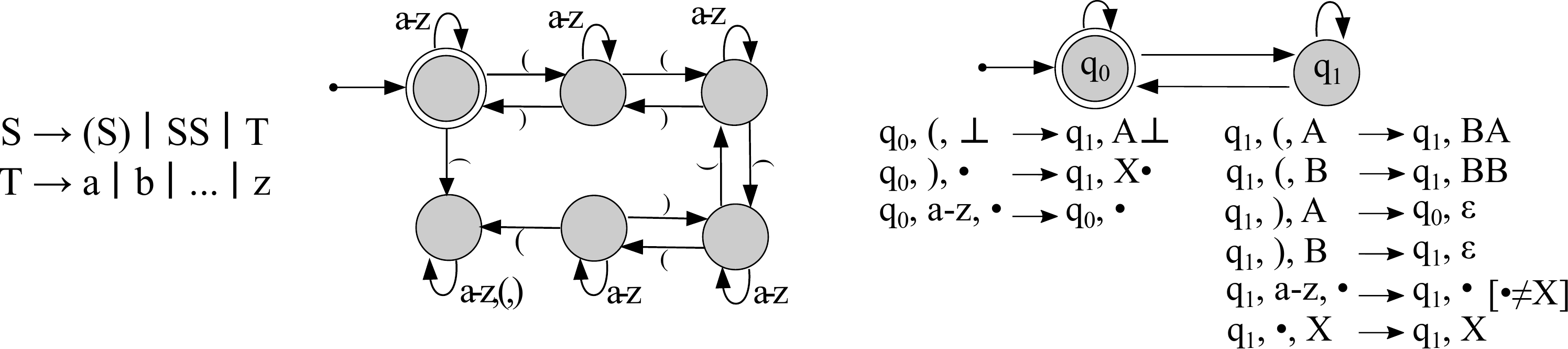} 
\caption{\label{fig:dfa-vs-pda} (Left) The context-free grammar for the language balanced parentheses (BP). (Center) A DFA that recognizes BP up to depth 4. (Right) A deterministic pushdown automaton (DPDA) that recognizes BP for all depths. The symbol $\bullet$ is a wildcard and stands for all possible tokens. The DPDA extrapolates to all sequences of BP, the DFA recognizes only those up to nesting depth 4.} %We aim to regularize standard RNNs so as to make them behave more like PDAs and less like DFAs.  }%
\end{figure*}

We propose state-regularized RNNs as a possible step towards addressing both of the aforementioned problems. State-regularized RNNs (\srrns) are a class of recurrent networks that utilize a stochastic state transition mechanism between cell applications.  The stochastic mechanism models a probabilistic state dynamics that lets the \srrns transition between a finite number of learnable states. The parameters of the stochastic mechanism are trained jointly with the parameters of the base RNNs. 

\srrns have several advantages over standard RNNs.  First, instead of having to apply post-training DFA extraction, \srrns determine their (probabilistic and deterministic) state transition behavior more directly. We propose a method that extracts DFAs truly representing the state transition behavior of the underlying RNNs. 
Second, we hypothesize that the frequently-observed poor extrapolation behavior of RNNs is caused by memorization with hidden states. It is known that RNNs -- even those with cell states or external memory -- tend to  memorize mainly with their hidden states and in an unstructured manner~\citep{strobelt2016visual,hao:2018}. We show that the state-regularization mechanism shifts representational power to memory components such as the cell state, resulting in improved extrapolation performance.

%State-regularization forces RNNs to operate less like DFAs with large state spaces and more like automata with external memory. 

We support our hypotheses through experiments both on synthetic and real-world datasets.  We explore the  improvement of the extrapolation capabilities of  \srrns and closely investigate their memorization behavior. For state-regularized LSTMs, for instance, we observe that memorization can be shifted entirely from the hidden state to the cell state. For text and visual data, state-regularization provides more intuitive interpretations of the RNNs' behavior. %We can also show that \srrns do not exhibit the drifting behavior of hidden states typical for standard RNNs on longer sequences. 

\section{Background}
\label{background}

%\subsection{Regular Languages and Finite State Machines}
We provide some background on deterministic finite automata (DFAs) and deterministic pushdown automata (DPDAs) for two reasons. First, one contribution of our work is a method for extracting DFAs from RNNs. Second, the state regularization we propose is intended to make RNNs behave more like DPDAs and less like DFAs by limiting their ability to memorize with hidden states.%, and, therefore, increasing their extrapolation abilities. 

A DFA is a state machine that  accepts or rejects sequences of tokens and produces one unique computation path for each input. 
Let $\Sigma^{*}$ be the language over the alphabet $\Sigma$ and let $\epsilon$ be the empty sequence.  A DFA over an alphabet (set of tokens) $\Sigma$ is a 5-tuple $(\calQ, \Sigma, \delta, q_0, F)$ consisting of
finite set of states $\calQ$; a finite set of input tokens $\Sigma$ called the input alphabet; a transition functions $\delta : \calQ \times \Sigma   \rightarrow \calQ$; a start state $q_0$; and a set of accept states $F \subseteq \calQ$.
A sequence $w$ is accepted by the DFA if the application of the transition function, starting with $q_0$, leads to an accepting state. 
Figure~\ref{fig:dfa-vs-pda}(center) depicts a DFA for the language of balanced parentheses (BP) up to depth 4. A language is regular if and only if it is accepted by a DFA.

A pushdown automata (PDA) is defined as a 7-tuple $(\calQ, \Sigma, \Gamma, \delta, q_0, \perp, F)$ consisting of 
a finite set of states $\calQ$; a finite set of input tokens $\Sigma$ called the input alphabet; a finite set of tokens $\Gamma$ called the stack alphabet; a transition function $\delta \subseteq \calQ \times (\Sigma \cup \epsilon) \times \Gamma \rightarrow \calQ \times \Gamma^{*}$; a start state $q_0$; the initial stack symbol $\perp$; and a set of accepting states $F\subseteq \calQ$. Computations of the PDA are applications of the transition relations. The computation starts in $q_0$ with the initial stack symbol $\perp$ on the stack and sequence $w$ as input. The pushdown automaton accepts $w$ if after reading $w$ the automaton reaches an accepting state. 
Figure~\ref{fig:dfa-vs-pda}(right) depicts a deterministic PDA for the language BP.

\section{State-Regularized Recurrent Networks}

The standard recurrence of an RNN is $\bfh_t = f\left(\bfh_{t-1},\bfc_{t-1},\bfx_{t}\right)$ where $\bfh_t$ is the hidden state vector at time $t$, $\bfc_t$ is the cell state at time $t$, and $\bfx_t$ is the input symbol at time $t$.  We refer to RNNs whose unrolled cells are only connected through the hidden output states $\bfh$ and no additional vectors such as the cell state, as \emph{memory-less} RNNs.  For instance, the family of \grus~\citep{Chung:2014} does not have cell states and, therefore, is memory-less. \lstms~\citep{Hochreiter:1997}, on the other hand, are not memory-less due to their cell state. 

A cell of a state-regularized RNN (\srrn) consist of two components. The first component, which we refer to as the \emph{recurrent component}, applies the function of a standard RNN cell
\begin{equation}
\bfu_{t} = f\left(\bfh_{t-1},\bfc_{t-1},\bfx_{t}\right).
\end{equation}
For the sake of completeness, we include the cell state $\bfc$ here, which is absent in memory-less RNNs. 

We propose a second component which we refer to as \emph{stochastic component}. The stochastic component is responsible for modeling the probabilistic state transitions  that let the RNN transition implicitly between a finite number of states. Let $d$ be the size of the hidden state vectors of the recurrent cells. Moreover, let $\Delta^{D} := \{ \bm{\lambda} \in \mathbb{R}_{+}^{D} : \parallel \bm{\lambda} \parallel = 1\}$ be the $(D-1)$ probability simplex.  The stochastic component maintains $k$ learnable centroids $\bfs_1$, ..., $\bfs_k$ of size $d$ which we often write as the column vectors of a matrix $\mathbf{S} \in \mathbb{R}^{d \times k}$. The weights of these centroids are global parameters shared among all cells. The stochastic component computes, at each time step $t$, a discrete probability distribution from the output $\bfu_t$ of the recurrent component  and the centroids of the stochastic component
\begin{equation}
\bm{\alpha} = g(\mathbf{S}, \bfu_t)  \mbox{ with }  \bm{\alpha} \in \Delta^{k} .  
\end{equation}
Crucially, instances of $g$ should be differentiable to facilitate end-to-end training. Typical instances of the function $g$ are based on the dot-product or the Euclidean distance, normalized into a probability distribution
%\vspace{-10mm}
\begin{align}
  \label{eqn-prob-attention}
    \alpha_i = & \frac{ \exp\left((\bfu_{t} \cdot \bfs_i)  / \tau\right) }{ \sum_{i=1}^{k} \exp\left((\bfu_{t} \cdot \bfs_i) / \tau\right)} \\
  \label{eqn-prob-eucledian}
    \alpha_i = & \frac{ \exp\left( -\parallel \bfu_{t}- \bfs_i\parallel  / \tau\right) }{ \sum_{i=1}^{k} \exp\left(-\parallel \bfu_{t}- \bfs_i\parallel / \tau\right)}
  \end{align}
Here, $\tau$ is a temperature parameter that can be used to anneal the probabilistic state transition behavior. The lower $\tau$ the more $\bm{\alpha}$ resembles the one-hot encoding of a centroid. The higher $\tau$ the more uniform is $\bm{\alpha}$. 
Equation~\ref{eqn-prob-attention} is reminiscent of the equations of attentive mechanisms~\citep{bahdanau2014neural,NIPS2017_7181}. Instead of attending to the hidden states, however, \srrns attend to the $k$ centroids to compute transition probabilities. Each $\alpha_i$ is the probability of the RNN to transition to centroid (state) $i$ given the vector $\bfu_t$ for which we write $p_{\bfu_t}(i) =  \alpha_i$.

\begin{figure}[htb]
%\vspace{-5mm}
\centering
\includegraphics[width=0.73\textwidth]{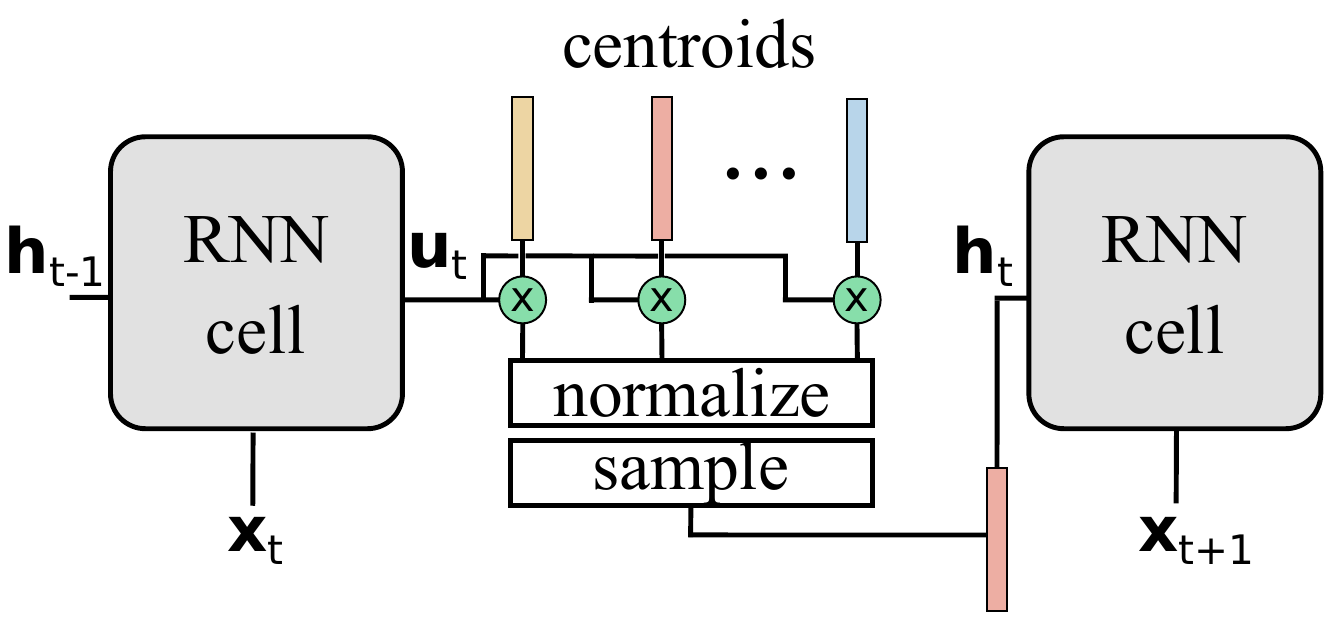}\\
\includegraphics[width=0.73\textwidth]{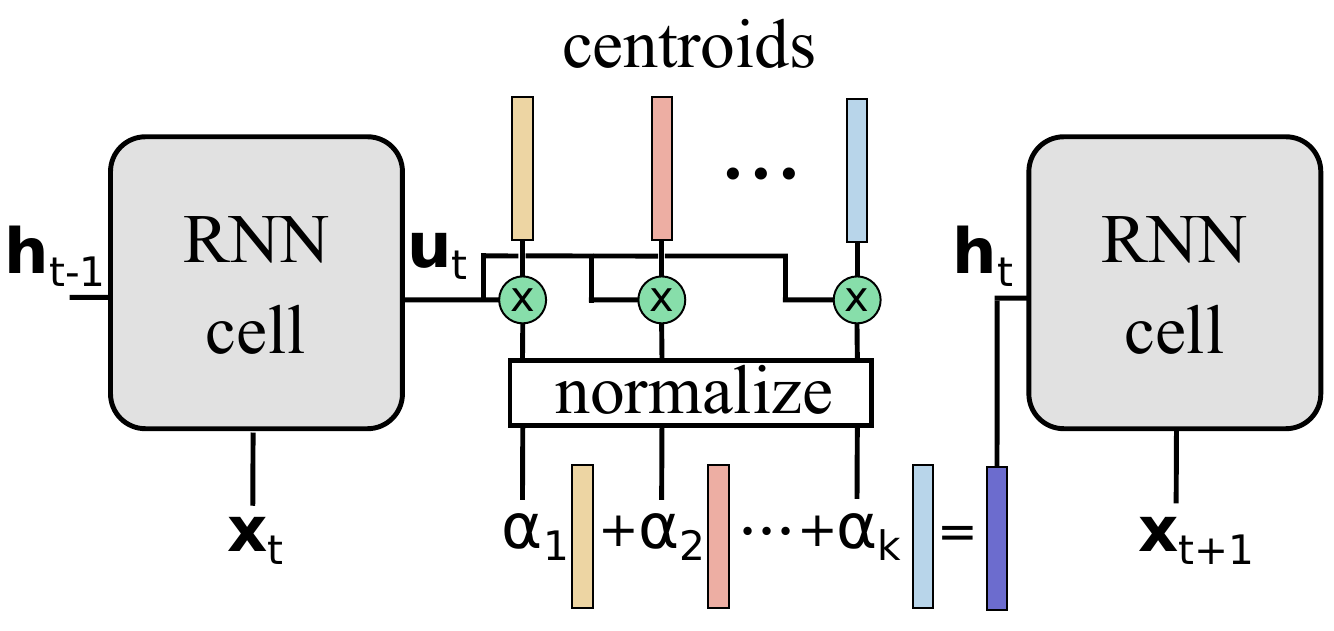}\\
\caption{\label{fig:srnn-example} Two possible instances of an \srrn corresponding to equations~\ref{eqn-prob-attention}\&\ref{eqn-transition-1} and ~\ref{eqn-prob-attention}\&\ref{eqn-transition-2}. }%
\vspace{-5mm}
\end{figure}

The state transition dynamics of an \srrn is that of a probabilistic finite state machine. At each time step, when being in state $\bfh_{t-1}$ and reading input symbol $\bfx_t$, the probability for transitioning to state $\bfs_i$ is $\alpha_i$. Hence, in its second phase the stochastic component computes the hidden state $\bfh_t$ at time step $t$ from the distribution $\bm{\alpha}$ and the matrix $\bfS$ with a (possibly stochastic) mapping $h: \Delta^{k}\times \mathbb{R}^{d\times k} \rightarrow \mathbb{R}^{d}$. Hence, $\bfh_t = h(\bm{\alpha},\bfS)$. An instance of  $h$ is to 
\begin{equation}
\label{eqn-transition-1}
\mbox{sample } j \sim p_{\bfu_t} \mbox{  and set } \bfh_t = \bfs_j.
\end{equation}
This renders the \srrn not end-to-end differentiable, however, and one has to use EM or reinforcement learning strategies which are often less stable and less efficient. 
A possible alternative is to set the hidden state $\bfh_t$ to be the probabilistic mixture of the centroids 
\begin{equation}
\label{eqn-transition-2}
\bfh_t = \sum_{i=1}^{k} \alpha_i \bfs_i.
\end{equation}
Every internal state $\bfh$ of the \srrn, therefore, is computed as a weighted sum $\bfh = \alpha_1 \bfs_1 + ... + \alpha_k \bfs_k$ of the centroids $\bfs_1, ..., \bfs_k$ with $\bm{\alpha} \in \Delta^{k}$. Here, $h$ is a smoothed variant of the function that computes a hard assignment to one of the centroids.  One can show that for $\tau \rightarrow 0$ the state dynamics based on equations (4) and (5) are identical and correspond to those of a DFA. Figure~\ref{fig:srnn-example} depicts two variants of the proposed \srrns.

Additional instances of $h$ are conceivable. For instance, one could, for every input sequence and the given current parameters of the \srrn, compute the most probable state sequence and then backpropagate based on a structured loss. Since finding these most probable sequences is possible with Viterbi type algorithms, one can apply a form of  differentiable dynamic programming~\citep{pmlr-v80-mensch18a}. The probabilistic state transitions of \srrns open up new possibilities for applying more complex differentiable  functions. We leave these considerations to future work. 
The probabilistic state transition mechanism is also applicable when RNNs have more than one hidden layer. In RNNs with $l>1$ hidden layers, every such layer can maintain its own centroids and stochastic component. In this case, a global state of the \srrn is an $l$-tuple, with the $l{\mbox{th}}$ argument of the tuple corresponding to the centroids of the $l{\mbox{th}}$ layer. 

%Before we explore its properties, let us take a step back and discuss the implications of using a stochastic component. 
Even though we have augmented the original RNN with additional learnable parameter vectors, we are actually constraining the \srrn to output hidden state vectors that are similar to the centroids. For lower temperatures and smaller values for $k$, the ability of the \srrn to memorize with its hidden states is increasingly impoverished. We argue that this behavior is beneficial for three reasons. First, it makes the extraction of interpretable DFAs from memory-less \srrns straight-forward. Instead of applying post-training DFA extraction as in previous work, we extract the true underlying DFA directly from the \srrn. Second, we hypothesize that overfitting in the context of RNNs is often caused by memorization via hidden states. Indeed, we show that regularizing the state space pushes representational power to memory components such as the cell state of an \lstm, resulting in improved extrapolation behavior. Third, the values of hidden states tend to increase in magnitude with the length of the input sequence, a behavior that has been termed \emph{drifting}~\citep{zeng1993learning}. The proposed state regularization stabilizes the hidden states for longer sequences.
\begin{figure*}[!htb]
\centering
\subfloat{{\includegraphics[width=0.74\textwidth,valign=b]{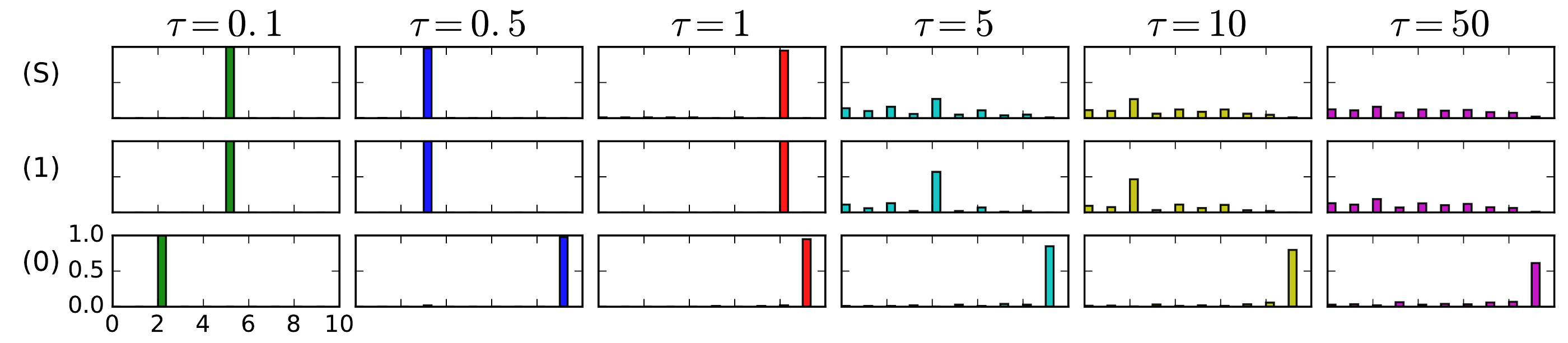} }}%
\hspace{1mm}
\subfloat{{\includegraphics[width=0.17\textwidth,valign=b]{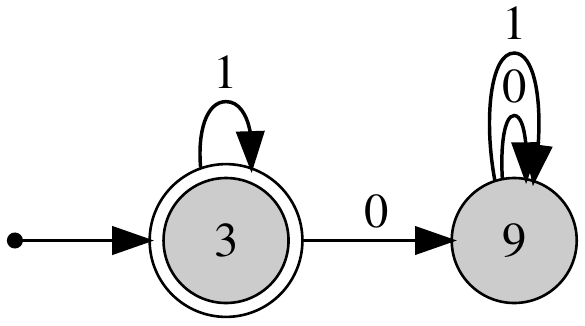} }}%
\vspace{-2mm}
\caption{\label{fig:example-tomita-probs} (Left) State transition probabilities for the \srgru learned from the data for the Tomita~1 grammar, for temperatures $\tau$ and input sequence $\mathtt{[10]}$. $\mathtt{S}$ is the start token. Centroids are listed on x-axis, probabilities on y-axis. Up to temperature $\tau = 1$ the behavior of the trained \srgrus is almost identical to that of a DFA. Despite the availability of $k=10$ centroids, the trained \srgrus use the minimal set of states for $\tau \leq 1$. (Right) The extracted DFA for Tomita grammar 1 and temperature $\tau = 0.5$. }%
\end{figure*}

First, let us explore some of the theoretical properties of the proposed mechanism. We show that the addition of the stochastic component, when capturing the complete information flow between cells as, for instance, in the case of GRUs, makes the resulting RNN's state transition behavior identical to that of a probabilistic finite state machine. 
\begin{theorem}
\label{theorem-pdfa}
The state transition behavior of a memory-less \srrn using equation~\ref{eqn-transition-1} is identical to that of a probabilistic finite automaton. 
\end{theorem}

\begin{theorem}
\label{theorem-dfa-equiv}
For $\tau \rightarrow 0$ the state transition behavior of a memory-less \srrn (using equations~\ref{eqn-transition-1} or \ref{eqn-transition-2}) is equivalent to that of a deterministic finite automaton. 

\end{theorem}
We can show that the lower the temperature the more memory-less RNNs operate like DFAs. The proofs of the theorems are part of the Supplementary Material. 

\subsection{Learning DFAs with State-Regularized RNNs}
\label{interpretable-rnn}

Extracting DFAs from RNNs is motivated by applications where a thorough understanding of learned neural models is required before deployment. \srrns maintain a set of learnable states and compute and explicitly follow state transition probabilities. It is possible, therefore, to extract finite-state transition functions that truly model the underlying state dynamics of the \srrn. The centroids do not have to be extracted from a clustering of a number of observed hidden states but can be read off of the trained model. This renders the extraction also more efficient. We adapt previous work~\citep{Schellhammer:1998,Wang:2007:nc} to construct the transition function of a \srrn. We begin with the start token of an input sequence, compute the transition probabilities $\bm{\alpha}$, and move the \srrn to the highest probability state. We continue this process until we have seen the last input token. By doing this, we get a count of transitions from every state $\bfs_i$ and input token $a\in\Sigma$ to the following states (including selfloops).
After obtaining the transition counts, we keep only the most frequent transitions and discard all other transitions. Due to space constraints, the pseudo-code of the extraction algorithm is listed in the Supplementary Material. 
As a corollary of Theorem~\ref{theorem-dfa-equiv} we have that, for $\tau \rightarrow 0$, the extracted transition function is identical to the transition function of the DFA learned by the \srrn. Figure~\ref{fig:example-tomita-probs} shows that for a wide range of temperatures (including the standard softmax temperature $\tau=1$) the transition behavior of a \srgru is identical to that of a DFA, a behavior we can show to be common when \srrns are trained on regular languages.

\begin{figure*}[t!]
\vspace{-5mm}
\centering
%\subfloat[]{{\includegraphics[width=0.15\textwidth,valign=b]{GRU_tomita_1_dfa}}}%
%\hspace{1mm}
\subfloat{{\includegraphics[width=0.2\textwidth,valign=b]{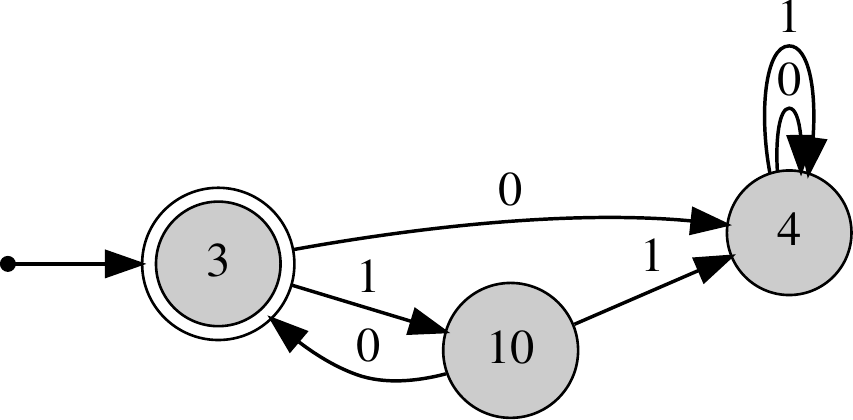} }}%
\hspace{2mm}
\subfloat{{\includegraphics[width=0.28\textwidth,valign=b]{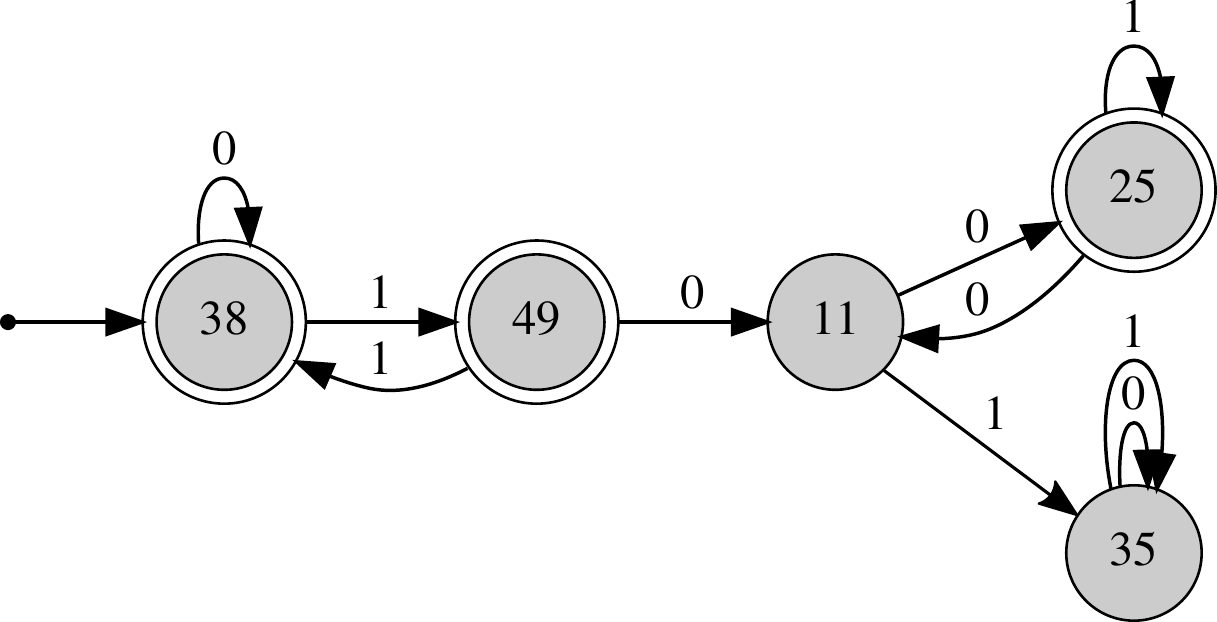} }}%
\hspace{2mm}
\subfloat{{\includegraphics[width=0.28\textwidth,valign=b]{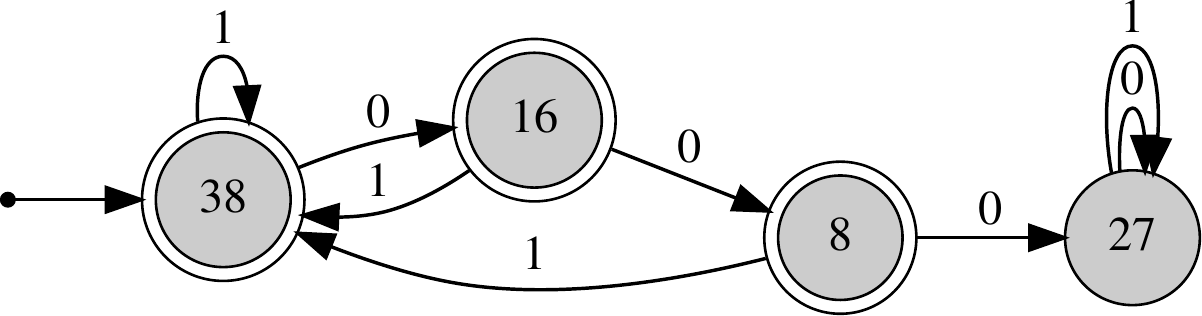} }}%
\caption{\label{fig:example-tomita-dfas} Extracted DFAs of the Tomita grammars 2-4. The state numbers correspond to the index of the learned \srgru centroids.}%
\end{figure*}

\subsection{Learning Nonregular Languages with State-Regularized LSTMs}
\label{state-regularization}
\begin{table*}[t!]
\begin{center}
\begin{small}
\begin{tabular}{lccc|ccc}
\toprule
Dataset & \multicolumn{3}{c|}{Large Dataset} & \multicolumn{3}{c}{Small Dataset} \\
Models & \lstm    & \srlstm   & \srlstmp   & \lstm    & \srlstm   & \srlstmp   \\ 
\midrule
$d\in{[}1, 10{]}$, $l\leq 100$  & 0.005   & 0.038     & \textbf{0.000}       & 0.068   & 0.037     & \textbf{0.017}       \\
$d\in{[}10, 20{]}$, $l\leq 100$ & 0.334   & 0.255     & \textbf{0.001}       & 0.472   & 0.347     & \textbf{0.189}       \\
$d\in{[}10, 20{]}$, $l\leq 200$ & 0.341   & 0.313     & \textbf{0.003}       & 0.479   & 0.352     & \textbf{0.196}       \\ 
$d=5$,  $l\leq 200$     & 0.002   & 0.044     & \textbf{0.000}       & 0.042   & 0.028     & \textbf{0.015}       \\
$d=10$, $l\leq 200$     & 0.207   & 0.227     & \textbf{0.004}       & 0.409   & 0.279     & \textbf{0.138}       \\
$d=20$, $l\leq 1000$    & 0.543   & 0.540     & \textbf{0.020}       & 0.519   & 0.508     & \textbf{0.380}      \\ 
\bottomrule
\end{tabular}
\end{small}
\end{center}
\vspace{-4mm}
\caption{\label{tab:bp}Error rates for the balanced parentheses (BP) test sets ($d$=depth, $l$=length, $k$=5 centroids, the training depth $\leq 5$).}
\end{table*}
For more complex languages such as context-free languages, RNNs that behave like DFAs generalize poorly to longer sequences. The DPDA shown in Figure~\ref{fig:dfa-vs-pda}, for instance, correctly recognizes the language of BP while the DFA only recognizes it up to nesting depth 4. 
We want to encourage RNNs with memory to behavor more like DPDAs and less like DFAs.  
The transition function $\delta$ of a DPDA takes (a) the current state, (b) the current top stack symbol, and (c) the current input symbol and maps these inputs to (1) a new state and (2) a replacement of the top stack symbol (see section~\ref{background}). Hence, to allow an \srrn such as the \srlstm to operate in a manner similar to a DPDA we need to give the RNNs access to these three inputs when deciding what to forget from and what to add to memory. Precisely this is accomplished for \lstms with peephole connections~\citep{Gers:2000}. The following additions to the functions defining the forget, input, and output gates include the cell state into the  \lstm's memory update decisions
\begin{alignat}{6}
\label{eqn-peephole}
& \textcolor{ashgrey}{\bff_t} & \textcolor{ashgrey}{=} & \ \textcolor{ashgrey}{\sigma \big(\bfW^{f}\bfx_t}  & \textcolor{ashgrey}{+} & \ \textcolor{ashgrey}{\bfR^{f}\bfh_{t-1}}  &  + & \ \textcolor{ceruleanblue}{\bfp^{f} \odot \bfc_{t-1}}  &  \textcolor{ashgrey}{+} & \textcolor{ashgrey}{\bfb^{f}} & \textcolor{ashgrey}{\big)}& ~~~~~~~~~~~~\\
& \textcolor{ashgrey}{\bfi_t} & \textcolor{ashgrey}{=} & \ \textcolor{ashgrey}{\sigma\big( \bfW^{i}\bfx_t}  & \textcolor{ashgrey}{+} & \ \textcolor{ashgrey}{\bfR^{i}\bfh_{t-1}}  & + & \ \textcolor{ceruleanblue}{\bfp^{i} \odot \bfc_{t-1}}  & \textcolor{ashgrey}{+} & \textcolor{ashgrey}{\bfb^{i}} & \textcolor{ashgrey}{\big)} & \\
& \textcolor{ashgrey}{\bfo_t} & \textcolor{ashgrey}{=} & \ \textcolor{ashgrey}{\sigma\big(\bfW^{o}\bfx_t} & \textcolor{ashgrey}{+} & \ \textcolor{ashgrey}{\bfR^{o}\bfh_{t-1}} &  + & \ \textcolor{ceruleanblue}{\bfp^{o} \odot \bfc_{t}}  & \textcolor{ashgrey}{+} & \textcolor{ashgrey}{\bfb^{o}} & \textcolor{ashgrey}{\big)}&
\end{alignat}
\color{black}
Here, $\bfh_{t-1}$ is the output of the previous cell's stochastic component; $\bfW$s and $\bfR$s are the matrices of the original LSTM; the $\bfp$s are the parameters of the peephole connections; and $\odot$ is the elementwise multiplication. 
We show empirically that the resulting \srlstmp operates like a DPDA, incorporating the current cell state when making decisions about changes to the next cell state.

\subsection{Practical Considerations}

Implementing \srrns only requires extending existing RNN cells with a stochastic component. We have found the use of  start and end tokens to be beneficial. The start token is used to transition the \srrn to a centroid representing the start state which then does not have to be fixed a priori. The end token is used to perform one more cell application but without applying the stochastic component before a classification layer. The end token lets the \srrn consider both the cell state and the hidden state to make the accept/reject decision. We find that a temperature of $\tau=1$ (standard softmax) and an initialization of the centroids with values sampled uniformly from $[-0.5,0.5]$ work well across different datasets. %We investigate the impact of the number of centroids more thoroughly in the experiments.  %We provide more experiment-specific details in later sections.

\begin{figure*}[!htb]
\vspace{0mm}
\begin{floatrow}
\capbtabbox{%

\begin{small}
\begin{tabular}{lcccc}
\toprule
Number of centroids & $k=2$   & $k=5$   & $k=10$  & $k=50$   \\ 
\midrule
$d\in {[}1, 10{]}$, $l\leq100$  & 0.019 & \textbf{0.017} & 0.021 & 0.034  \\
$d\in {[}10, 20{]}$, $l\leq100$ & \textbf{0.096} & 0.189 & 0.205 & 0.192  \\
$d\in {[}10, 20{]}$, $l\leq200$ & \textbf{0.097} & 0.196 & 0.213 & 0.191  \\  
$d=5$,  $l\leq200$     & 0.014 & 0.015 & \textbf{0.012} & 0.047  \\
$d=10$, $l\leq200$     & \textbf{0.038} & 0.138 & 0.154 & 0.128  \\
$d=20$, $l\leq1000$    & 0.399 & \textbf{0.380} & 0.432 & 0.410 \\  
\bottomrule
\end{tabular}
\end{small}
\vspace{2mm}
}{%
 \vspace{-5mm}
  \caption{\label{tab:bp_k}Error rates of the \srlstmp on the small BP test data for various numbers of centroids $k$ ($d$=depth, $l$=length).}%
}
\ffigbox[5cm]{%
\includegraphics[width=0.3\textwidth]{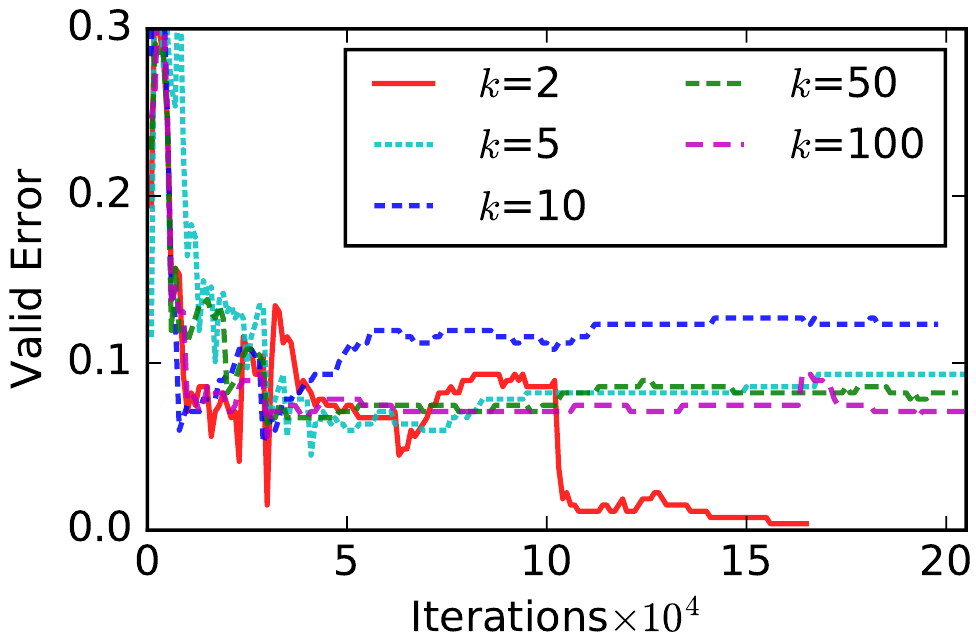}%
}{%
 \vspace{-6mm}
  \caption{\label{fig:error-small-bp-k} \srlstmp error curves on the small BP validation data. }%
}
\end{floatrow}
\end{figure*}

\begin{figure*}
\vspace{-5mm}
\subfloat[$\bfh_t$ of the \lstm]{{\includegraphics[width=0.275\textwidth,valign=b]{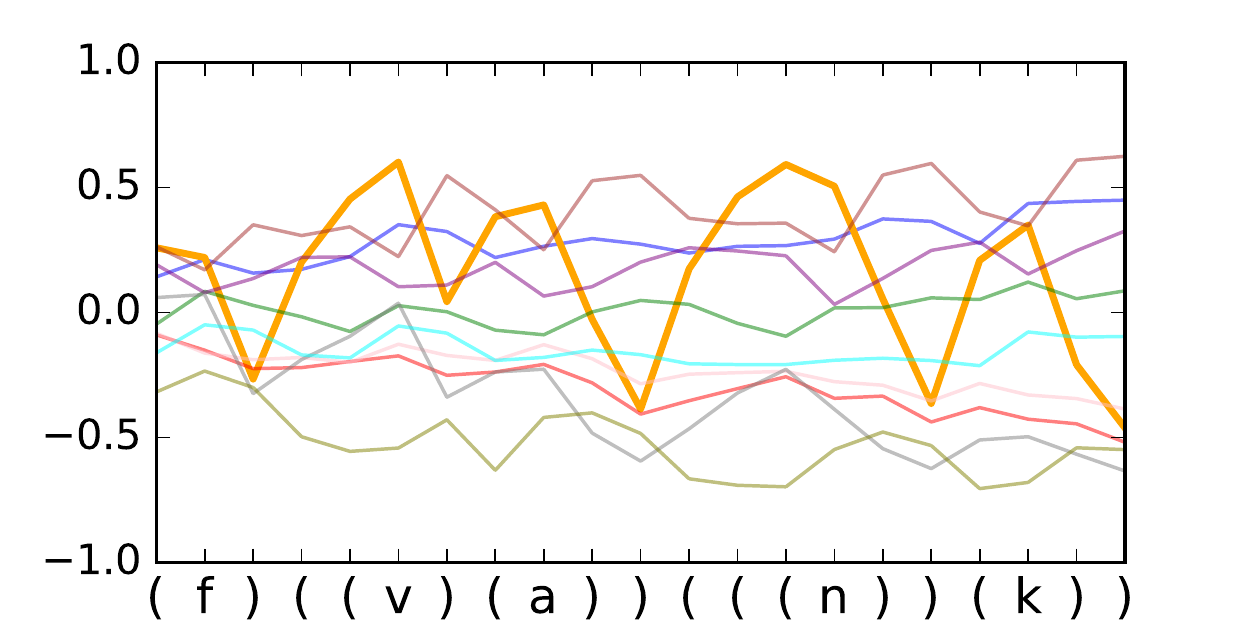} } \hspace{-6.25mm}}% 
\subfloat[ $\bfc_t$ of the \lstm]{{\includegraphics[width=0.275\textwidth,valign=b]{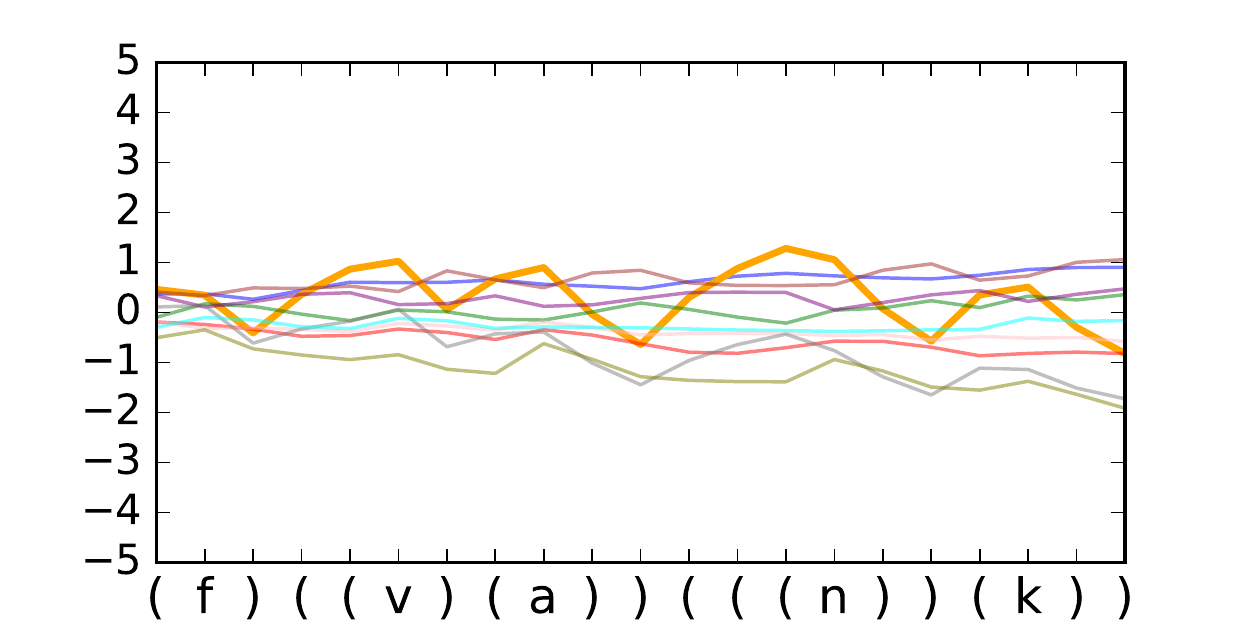} }  \hspace{-6.25mm}} 
\subfloat[ $\bfh_t$ of the \srlstmp]{{\includegraphics[width=0.275\textwidth,valign=b]{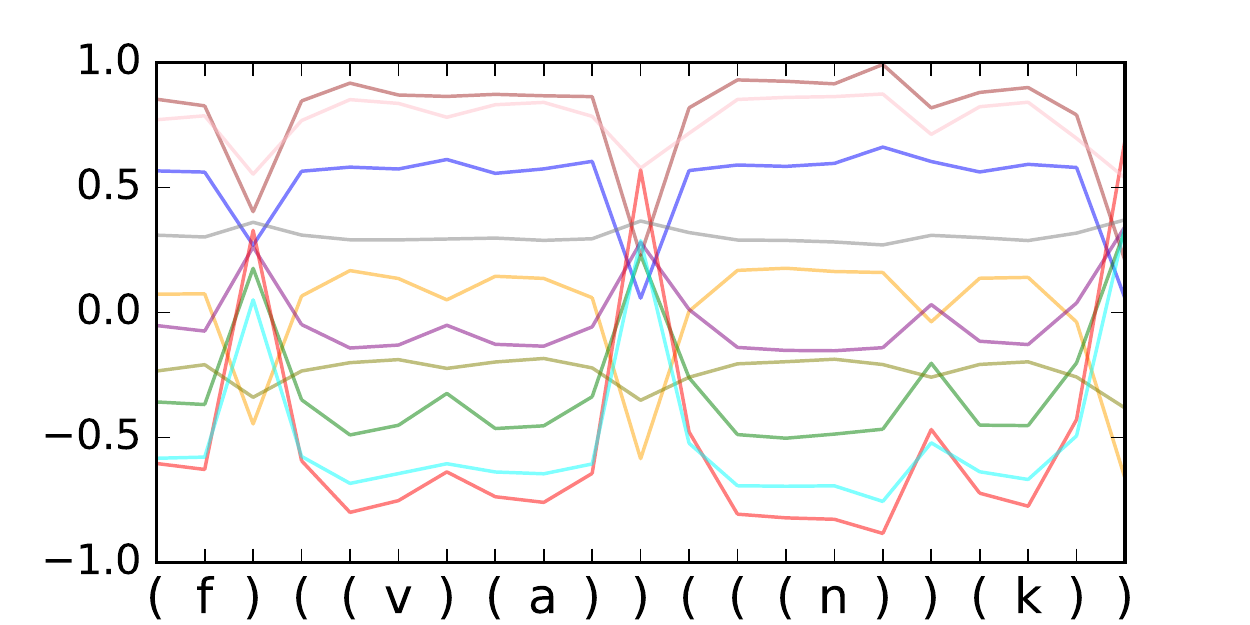} }   \hspace{-6.25mm}}% 
\subfloat[ $\bfc_t$ of the \srlstmp]{{\includegraphics[width=0.275\textwidth,valign=b]{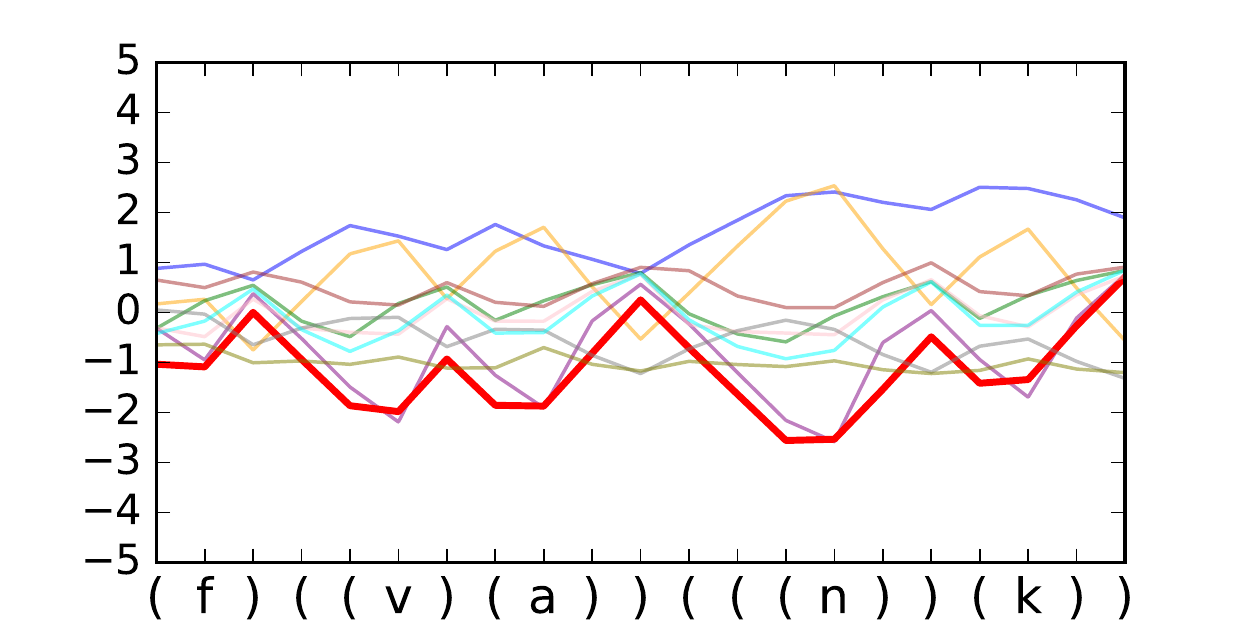} }  \hspace{-6.5mm}}% 
\caption{\label{fig:bp_vis} Visualization of hidden state $\bfh_t$ and cell state $\bfc_t$ of the \lstm and the \srlstmp for a specific input sequence from BP.  Each color corresponds to one of 10 hidden units. The \lstm memorizes the number of open parentheses both in the hidden and to a lesser extent in the cell state (bold yellow lines). The memorization is not accomplished with saturated gate outputs and a drift is observable for both vectors. The \srlstmp maintains two distinct hidden states (accept and reject) and does not visibly memorize counts through its hidden states. The cell state is used to cleanly memorize the number of open parentheses (bold red line) with saturated gate outputs ($\pm1$). For \srlstmp, a state vector drift is not observable; solutions with less drift generalize better~\cite{gers2001lstm}.}%
\vspace{-2mm}
\end{figure*}

\section{Experiments}

We conduct three types of experiments to investigate our hypotheses. First, we apply a simple algorithm for extracting DFAs and assess to what extent the true DFAs can be recovered from input data. Second, we compare the behavior of \lstms and state-regularized \lstm on nonregular languages such as the languages of balanced parentheses and palindromes. Third, we investigate the performance of state-regularized \lstms on non-synthetic datasets. 

Unless otherwise indicated we always (a) use single-layer RNNs, (b) learn an embedding for input tokens before feeding it to the RNNs, (c) apply \textsc{RMSprop} with a learning rate of $0.01$ and momentum of $0.9$; (d) do not use dropout~\cite{srivastava2014dropout} or batch normalization~\cite{cooijmans2016recurrent} of any kind; and (e) use state-regularized RNNs based on equations~\ref{eqn-prob-attention}\&\ref{eqn-transition-2} with a temperature of $\tau=1$ (standard softmax). We include more experimental results and the code in the Supplementary Material.

\subsection{Regular Languages and DFA Extraction}

We evaluate the DFA extraction algorithm for \srrns on RNNs trained on the Tomita grammars~\citep{tomita:cogsci82} which have been used as
benchmarks in previous  work \citep{Wang:2007:nc,pmlr-v80-weiss18a}. We use available code~\citep{pmlr-v80-weiss18a} to generate training and test data for the regular languages. We first trained a single-layer \gru with $100$ units on the data. We use GRUs since they are memory-less and, hence, Theorem~\ref{theorem-dfa-equiv} applies. Whenever the \gru converged within 1 hour to a training accuracy of $100\%$, we also trained a \srgru based on equations~\ref{eqn-prob-attention}\&\ref{eqn-transition-2} with $k=50$ and $\tau=1$. This was the case for the grammars 1-4 and 7. The difference in time to convergence between the vanilla \gru and the \srgru was negligible. We applied the transition function extraction outlined in section~\ref{interpretable-rnn}. In all cases, we could recover the minimal and correct DFA corresponding to the grammars. Figure~\ref{fig:example-tomita-dfas} depicts the DFAs for grammars 2-4 extracted by our approach. Remarkably, even though we provide more centroids (possible states; here $k=50$) the \srgru only utilizes the required minimal number of states for each of the grammars.  Figure~\ref{fig:example-tomita-probs} visualizes the transition probabilities for different temperatures and $k=10$ for grammar 1. The numbers on the states  correspond directly to the centroid numbers of the learned \srgru. One can observe that the probabilities are spiky, causing the \srgru to behave like a DFA for $\tau\leq1$. 

\subsection{Nonregular Languages}

%\subsubsection{Balanced Parentheses}

We conducted experiments on nonregular languages where external memorization is required. We wanted to investigate whether our hypothesis that \srlstm behave more like DPDAs and, therefore, extrapolate to longer sequences, is correct.  To that end, we used the context-free language ``balanced parentheses" (BP; see Figure~\ref{fig:dfa-vs-pda}(left)) over the alphabet $\Sigma=\{a, ..., z,(,)\}$,  used in previous work~\citep{pmlr-v80-weiss18a}. We created two datasets for BP. A large one with 22,286 training sequences (positive: 13,025; negative: 9,261) and 6,704 validation sequences (positive: 3,582; negative: 3,122). The small dataset consists of 1,008 training sequences (positive: 601; negative: 407), and 268 validation sequences (positive: 142; negative: 126). The training sequences have nesting depths $d \in [1, 5]$ and the validation sequences $d \in [6,10]$. We trained the \lstm and the \srrns using curriculum learning as in previous work \citep{Zaremba:2014,pmlr-v80-weiss18a} and using the validation error as stopping criterion. We then applied the trained models to unseen sequences.   Table \ref{tab:bp} lists the results on 1,000 test sequences each with the respective depths and lengths. The results show that both \srlstm and \srlstmps extrapolate better to longer sequences and sequences with deeper nesting. Moreover, the \srlstmp performs almost perfectly on the large data indicating that peephole connections are indeed beneficial.

To explore the effect of the hyperparameter $k$, that is, the number of centroids of the \srrns, we ran experiments on the small BP dataset varying $k$ and keeping everything else the same. Table~\ref{tab:bp_k} lists the error rates and Figure~\ref{fig:error-small-bp-k} the error curves on the validation data for the \srlstmp and different values of $k$. While two centroids ($k=2$) result in the best error rates for most sequence types, the differences are not very pronounced. This indicates that the \srlstmp is robust to changes in the hyperparameter $k$. A close inspection of the transition probabilities reveals that the \srlstmp mostly utilizes two states, independent of the value of $k$. These two states are used as accept and reject states. 
These results show that \srrns generalize tend to utilize a minimal set states similar to DPDAs. 

A major hypothesis of ours is that the state-regularization encourages RNNs to operate more like DPDAs. To explore this hypothesis, we trained an \srlstmp with $10$ units on the BP data and visualized both the hidden state $\bfh_t$ and the cell state $\bfc_t$ for various input sequences. Similar state visualizations have been used in previous work~\citep{strobelt2016visual,Weiss:2018-power}.
Figure~\ref{fig:bp_vis} plots the hidden and cell states for a specific input, where each color corresponds to a dimension in the respective state vectors. As hypothesized, the \lstm relies primarily on its hidden states for memorization. The \srlstmp, on the other hand, does not use its hidden states for memorization. Instead it utilizes two main states (accept and reject) and memorizes the nesting depth cleanly in the cell state. The visualization also shows a drifting behavior for the \lstm, in line with observations made for first-generation RNNs~\citep{zeng1993learning}. Drifting is not observable for the \srlstmp. 

\begin{figure*}[!htb]
%\vspace{-5mm}
\begin{floatrow}
\capbtabbox{%

\begin{small}

\begin{tabular}{p{2cm}p{0.52cm}p{0.52cm}p{0.52cm}}
\toprule
Max Length  &    100  & 200 & 500 \\
\midrule
%length  & & & \\  \hline \hline
\lstm &  31.2 & 42.0 & 47.7 \\
\lstmp &  28.4 & 36.2 & 41.5 \\
\srlstm &  28.0 & 36.0 & 44.6 \\
\srlstmp &  \textbf{10.5} & \textbf{16.7} & \textbf{29.8}  \\ 
\bottomrule
\end{tabular}
\end{small}
\vspace{-4mm}
}{%
  \caption{\label{tab:palindromes} Error rates in $\%$ on sequences of varying lengths from the Palindrome test set.}%
}
\capbtabbox{%
\hspace{-6mm}
\begin{small}
\begin{tabular}{p{1.85cm}p{0.72cm}p{0.72cm}p{0.72cm}p{0.72cm}p{0.72cm}p{0.72cm}p{0.72cm}p{0.72cm}}
\toprule
Model       & \multicolumn{2}{c}{\lstm}        & \multicolumn{2}{c}{\srlstm}      & \multicolumn{2}{c}{\lstmp}                      & \multicolumn{2}{c}{\srlstmp}               \\ 
Length      & $100$ & $200$ & $100$ & $200$ & $100$ & $200$ & $100$                      & $200$ \\ 
\midrule
train error & 29.23                  & 32.77                  & 27.17                  & 31.72                  &25.50                & 29.05                &  \textbf{23.30}                                       &  \textbf{24.67}                  \\
test error  & 29.96                  & 33.19                  & 28.82                  & 32.88                  & 28.02                  & 30.12                  & \textbf{26.04}                                       & \textbf{26.21}              \\
time (epoch)        & \multicolumn{2}{c}{400.29s}               & \multicolumn{2}{c}{429.15s}               & \multicolumn{2}{c}{410.29s}                            & \multicolumn{2}{l}{466.10s}                             \\ 
\bottomrule     
\end{tabular}
\end{small}
}{%
\vspace{-4mm}
 \caption{\label{tab:imdb_small}Error rates (on training and test splits of the IMDB data) in $\%$ and averaged training time in seconds when training only on truncated sequences of length $10$.}%
}
\end{floatrow}
\end{figure*}

\begin{figure*}[!htp]
\begin{floatrow}
\capbtabbox{%
\begin{small}
\begin{tabular}{p{6.5cm}c}
\toprule
Methods                                                              & Error \\ 
\midrule
\multicolumn{2}{c}{\textbf{use additional unlabeled data}}  \\
Full+unlabelled+BoW~(\citeauthor{maas-EtAl:2011:ACL-HLT2011})   & 11.1         \\
LM-LSTM+unlabelled~(\citeauthor{dai2015semi})              & 7.6            \\
SA-LSTM+unlabelled~(\citeauthor{dai2015semi})           & 7.2             \\  
\midrule
\multicolumn{2}{c}{\textbf{do not use additional unlabeled data}}  \\
seq2-bown-CNN~(\citeauthor{johnson2014effective})     & 14.7         \\
WRRBM+BoW(bnc)~(\citeauthor{dahl2012training})     & 10.8          \\
JumpLSTM~(\citeauthor{yu2017learning})       & 10.6         \\ 

\midrule
\lstm & 10.1  \\
\lstmp & 10.3  \\
\srlstm($k=10$) & 9.4  \\
\srlstmp($k=10$) & \textbf{9.2} \\
\srlstmp($k=50$) & 9.8  \\

\bottomrule
\end{tabular}
\end{small}
}{%
\vspace{-4mm}
  \caption{\label{tab:imdb} Test error rates (\%) on IMDB.}%
}
%\hspace{-4mm}

\capbtabbox{%

\caption{Test accuracy on pixel-by-pixel MNIST (\%)}
\label{tab:mnist_comparison}
\begin{small}
\begin{tabular}{p{5cm}c} 
\toprule
Methods& Accuracy  \\ 
\midrule
 %        & Normal  \\ \hline\hline
IRNN~(\citeauthor{le2015simple})            & 97.0      \\
URNN~(\citeauthor{arjovsky2016unitary})          & 95.1      \\
Full URNN~(\citeauthor{NIPS2016_6327})		&97.5		   \\ 
sTANH-RNN ~(\citeauthor{zhang2016architectural})   & 98.1 \\
RWA~(\citeauthor{ostmeyer2017machine})                 & 98.1      \\ 
Skip LSTM~(\citeauthor{campos2017skip})                    & 97.3    \\
r-LSTM Full BP~(\citeauthor{trinh2018learning})                    & 98.4    \\
BN-LSTM~(\citeauthor{cooijmans2016recurrent})                  & 99.0    \\
Dilated GRU~(\citeauthor{chang2017dilated}) & \textbf{99.2}    \\
\midrule
\lstm           & 97.7        \\
\lstmp           & 98.5        \\
\srlstm $(k=100)$ &98.6  \\
\srlstmp $(k=100)$ &\textbf{99.2}  \\
\srlstmp $(k=50)$ &98.6  \\
%\srlstmp $(k=10)$ &98.1    \\ 
\bottomrule
\end{tabular}
\end{small}
}{%
\vspace{-4mm}
  \caption{\label{tab:MNIST} Test accuracy (\%) on sequential MNIST.}%
}
\end{floatrow}
\end{figure*}

We also performed experiments for the nonregular language $ww^{-1}$ (Palindromes)~\cite{schmidhuber2002learning} over the alphabet $\Sigma = \{a, ..., z\}$. We follow the same experiment setup as for BP and include the details in the Supplementary Material. The results of Table \ref{tab:palindromes} add evidence for the improved generalization and memorization behavior of state-regularized LSTMs over vanilla LSTMs (with peepholes). Additional experimental results on the Copying Memory task are provided in the Supplementary Material. 

\subsection{Sentiment Analysis and Pixel-by-Pixel MNIST}

We evaluated state-regularized LSTMs on the IMDB review dataset~\citep{maas-EtAl:2011:ACL-HLT2011}.
%\footnote{\url{http://ai.stanford.edu/~amaas/data/sentiment/}}.
It consists of 100k movie reviews (25k training, 25k test, and 50k unlabeled).  We used only the labeled training and test reviews. Each review is labeled as \emph{positive} or \emph{negative}. To investigate the models' extrapolation capabilities, we trained all models on truncated sequences of up to length $10$ but tested on longer sequences (length 100 and 200). The results listed in Table \ref{tab:imdb_small} show the improvement in extrapolation performance of the \srlstmp. The table also lists the training time per epoch, indicating that the overhead compared to the standard \lstm (or with peepholes \lstmp) is modest. Table \ref{tab:imdb} lists the results when training without truncating sequences. The \srlstmp is competitive with state of the art methods that also do not use the unlabeled data.

We also explored the impact of state-regularization on pixel-by-pixel MNIST~\citep{lecun1998gradient,le2015simple}. Here, the 784 pixels of MNIST images are fed to RNNs one by one  for classification. This requires the modeling of long-term dependencies.   Table  \ref{tab:MNIST} shows the results on the normal and permuted versions of pixel-by-pixel MNIST. The classification function has the final hidden and cell state as input. Our (state-regularized) LSTMs do not use dropout, batch normalization, sophisticated weight-initialization, and are based on a simple single-layer LSTM. We can observe that \srlstmps achieve competitive results, outperforming the vanilla LSTM and LSTM-P. The experiments on the Language Modeling is in the Supplementary Material.

\begin{figure*}[!htb]
%\vspace{-2mm}
\begin{floatrow}
\capbtabbox{%
\small
\begin{tabular}{|ll|}
\hline
cent. & words with top-$4$ highest transition probabilities                                                                                                   \\ \hline\hline
1 &  but (0.97)    hadn (0.905)    college (0.87)    even (0.853)                                    \\
2 &  not (0.997)    or (0.997)    italian (0.995)    never (0.993)                                      \\
3 & {\color[HTML]{036400} loved (0.998)    definitely (0.996)    8 (0.993)    realistic (0.992)   } \\
4  & {\color[HTML]{FD6864} no (1.0)    worst (1.0)    terrible (1.0)    poorly (1.0)     }                                      \\ \hline
\end{tabular}
\vspace{-3mm}
}{%
 \caption{\label{tab:imdb_vis_5} The learned centroids and their prototypical words with the top-4 highest transition probabilities. This interprets the \srlstmp model with centroids. The $3^{rd}$ ($4^{th}$) centroid is ``positive'' (``negative'').}
}
%\hspace{-2mm}
\ffigbox[7.5cm]{%
\includegraphics[width=0.44\textwidth]{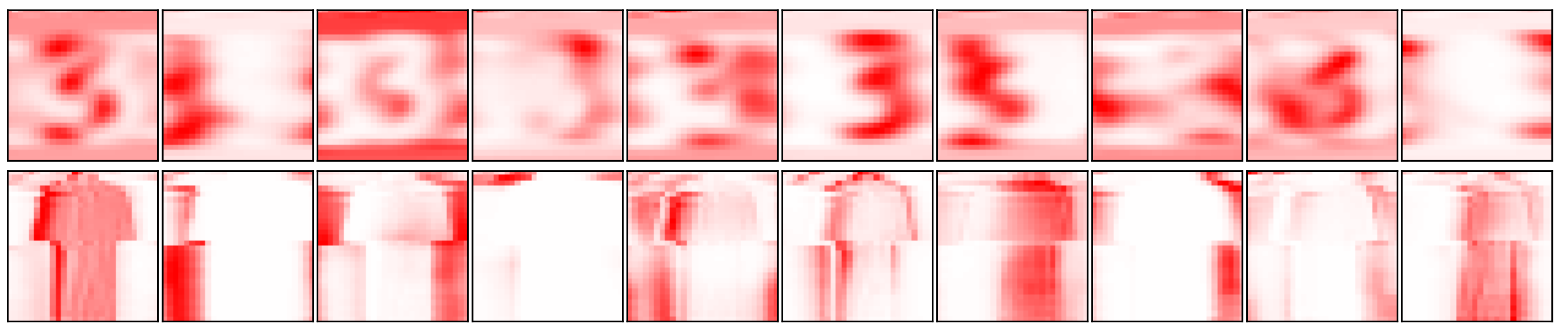} %
%\text{\small{ $0~~1~~~~2~~3~~~~4~~~~~~5~~~~~~~~~~6~~7~~~8~~~9~~~~~~~~~~~~~~~$}}
}{%
  \vspace{-5.5mm}
  \caption{\label{fig:explanations_mnist} The prototypes of generated by learned \srlstmp ($k$$=$$10$) centroids for ``3'' in MNIST (top) and ``T-shirt'' in Fashion-MNIST (bottom)~\cite{xiao2017fashion}. }%
}
\end{floatrow}
\end{figure*}
\subsection{Interpretability and Explainability}

State-regularization provides new ways to interpret the working of RNNs. Since \srrns have a finite set of states, we can use the observed transition probabilities to visualize their behavior. For instance, to generate prototypes for the \srrns we can select, for each state $i$, the input tokens that have the highest average probability leading to state $i$. For the IMDB reviews, these are the top probability words leading to each of the states. For pixel-by-pixel MNIST, these are the top probabilities of each pixel leading to the states. 
Table \ref{tab:imdb_vis_5} lists, for each state (centroid), the word with the top transition probabilities leading to this state. Figure \ref{fig:explanations_mnist} shows the prototypes associated with digits ``3'' of MNIST and ``T-shirt'' of Fashion-MNIST. In contrast to previous methods~\citep{berkes2006analysis,nguyen2016synthesizing}, the prototypes are directly generated with the centroids of the trained \srrns and the input token embeddings and hidden states do not have to be similar. For more evidence of generated interpretations and explanations, we refer to the Supplementary Material.% For explainability,  please see the Appendix for evidences. 

\section{Related Work}
\label{rel_work}

RNNs are powerful learning machines. Siegelmann and Sontag \citep{siegelmann1992computational, siegelmann1994analog, siegelmann2012neural}, for instance, proved that a variant of Elman-RNNs~\citep{elman1990finding} can simulate a Turing machine. Recent work considers the more practical situation where RNNs have  finite precision and linear computation time in their input length~\citep{Weiss:2018-power}.

Extracting DFAs from RNNs goes back to work on first-generation RNNs in the 1990s\citep{giles1991second,zeng1993learning}. These methods perform a clustering of hidden states after the RNNs are trained~\citep{wang2018comparison,frasconi1994approach,giles1991second}. Recent work introduced more sophisticated learning approaches to extract DFAs from LSTMs and GRUs~\citep{pmlr-v80-weiss18a}. The latter methods tend to be more successful in finding DFAs behaving similar to the RNN. In contrast to all existing methods, \srrns learn an explicit set of states which facilitates the extraction of DFAs from memory-less \srrns exactly modeling their state transition dynamics. %There is also previous work on visualizing the hidden and cell states of RNNs\citep{strobelt2016visual}.
A different line of work attempt to learn more interpretable RNNs~\cite{foerster2017input}, or rule-based classifiers from RNNs~\citep{murdoch2017automatic}.

There is a large body of work on regularization techniques for RNNs. Most of these adapt regularization approaches developed for feed-forward networks to the recurrent setting. Representative instances are dropout regularization~\citep{zaremba2014recurrent}, variational dropout~\citep{gal2016theoretically}, weight-dropped LSTMs~\citep{merity2017regularizing}, and noise injection \citep{dieng2018noisin}.
Two approaches that can improve convergence and generalization capabilities are batch normalization~\citep{cooijmans2016recurrent} and weight initialization strategies~\citep{le2015simple} for RNNs.
The work most similar to \srrns are self-clustering RNNs~\citep{zeng1993learning}. These RNNs learn discretized states, that is, binary valued hidden state vectors, and show that these networks generalize better to longer input sequences. Contrary to self-clustering RNNs, we propose an end-to-end differentiable probabilistic state transition mechanism between cell applications. 

Stochastic RNNs are a class of generative recurrent models for sequence data~\citep{bayer2014learning,Fraccaro:2016,Goyal:2017}. They model uncertainty in the hidden states of an RNN by introducing latent variables. %A core assumption of this line of work is that RNNs are good at modeling long-term dependencies. 
Contrary to \srrns, stochastic RNNs do not model probabilistic state transition dynamics. Hence, they do not address the problem of overfitting through hidden state memorization and improvements to DFA extraction.

There are proposals for extending RNNs with various types of external memory. Representative examples are the neural Turing machine~\citep{graves2014neural}, improvements thereof~\citep{Graves:2016b}, memory network~\cite{WestonCB14}, associative LSTM~\citep{danihelka2016associative}, and RNNs augmented with neural stacks, queues, and deques~\citep{grefenstette2015learning}. Contrary to these proposals, we do not augment RNNs with differentiable data structures but regularize RNNs to make better use of existing memory components such as the cell state. We hope, however, that differentiable neural computers could benefit from state-regularization. 

\section{Discussion and Conclusion}
State-regularization provides new mechanisms for understanding the workings of RNNs. Inspired by recent DFA  extraction work~\citep{pmlr-v80-weiss18a}, our work simplifies the extraction approach by directly learning a finite set of states and an interpretable state transition dynamics. Even on realistic tasks such as sentiment analysis, exploring the learned centroids and the transition behavior of \srrns makes for more interpretable RNN models whithout sacrificing accuracy: a single-layer \srrns is competitive with state-of-the-art methods. 
The purpose of our work is not to surpass all existing state of the art methods but to gain a deeper understanding of the dynamics of RNNs.

%\section{Conclusion}

State-regularized RNNs operate more like automata with external memory and less like DFAs. This results in a markedly improved extrapolation behavior on several datasets. We do not claim, however, that \srrns are a panacea for all problems associated with RNNs. For instance, we could not observe an improved convergence of \srrns. Sometimes \srrns converged faster, sometimes vanilla RNNs. While we have mentioned that the computational overhead of \srrns is modest, it still exists, and this might exacerbate the problem that RNNs often take a long to be trained and tuned. We plan to investigate variants of state regularization and the ways in which it could improve differentiable computers with RNN controllers.

\nocite{langley00}
\bibliography{icml_reference}
\bibliographystyle{icml2019}

\clearpage
\begin{appendices}

\section{Proofs of Theorems \ref{theorem-pdfa} and \ref{theorem-dfa-equiv}}

\subsection{Theorem \ref{theorem-pdfa}}

\textbf{The state transition behavior of a memory-less \srrn using equation~\ref{eqn-transition-1} is identical to that of a probabilistic finite automaton.}

\begin{figure*}
\begin{align}
\alpha_i = & \frac{ \exp\left( b_i  / \tau\right) }{ \exp(\left(b_M-\epsilon_1) / \tau\right) + ... + \exp\left(b_M / \tau\right) + ... + \exp(\left(b_M-\epsilon_k) / \tau\right)}
\label{poorf_1} \\
& =\frac{ \exp\left( b_i  / \tau\right) }{ \exp(b_M/\tau)\exp(\epsilon_1/ \tau)^{-1} + ... + \exp(b_M / \tau) + ... + \exp(b_M/\tau)\exp(\epsilon_k/\tau)^{-1}} 
\label{poorf_2}  \\
& =\frac{ \exp\left( b_i  / \tau\right) }{ \exp(b_M/\tau)\left[ \exp(\epsilon_1/ \tau)^{-1} + ... + 1 + ... + \exp(\epsilon_k/\tau)^{-1}\right]}
\label{poorf_3} .
\end{align}
\end{figure*}

\begin{proof}
The state transition function $\delta$ of a probabilistic finite state machine is identical to that of a finite deterministic automaton (see section~\ref{background})  with the exception that it returns a probability distribution over states. For every state $q$ and every input token $a$ the transition mapping $\delta$ returns a probability distribution $\bm{\alpha} = (\alpha_1, ..., \alpha_k)$ that assigns a fixed probability to each possible state $q \in \mathcal{Q}$ with $|\mathcal{Q}|=k$. The automaton transitions to the next state according to this distribution. Since by assumption the \srrn is using equation~\ref{eqn-transition-1}, we only have to show that the probability distribution over states computed by the stochastic component of a memory-less \srrn is identical for every state $q$ and every input token $a$ irrespective of the previous input sequence and corresponding state transition history. 

More formally, for every pair of input token sequences $\mathbf{a}_1$ and $\mathbf{a}_2$ with corresponding pair of resulting state sequences $\mathbf{q}_1 = (q_{i_1}, ..., q_{i_n}, q)$ and $\mathbf{q}_2 = (q_{j_1}, ..., q_{j_m}, q)$ in the memory-less \srrn, we have to prove, for every token $a \in \Sigma$, that $\bm{\alpha}_1$ and $\bm{\alpha}_2$, the probability distributions over the states returned by the stochastic component for state $q$ and input token $a$, are  identical. Now, since the RNN is, by assumption, memory-less, we have for both $\mathbf{a}_1, \mathbf{q}_1$ and $\mathbf{a}_2, \mathbf{q}_2$ that the only inputs to the RNN cell are exactly the centroid $\mathbf{s}_{q}$ corresponding to state $q$ and the vector representation of token $a$. Hence, under the assumption that the parameter weights of the RNN are the same for both state sequences $\mathbf{q}_1$ and $\mathbf{q}_2$, we have that the output $\mathbf{u}$ of the recurrent component (the base RNN cell) is identical for $\mathbf{q}_1$ and $\mathbf{q}_2$. Finally, since by assumption the centroids $\mathbf{s}_1, ..., \mathbf{s}_k$ are fixed, we have that the returned probability distributions $\bm{\alpha}_1$ and  $\bm{\alpha}_2$ are identical. Hence, the memory-less \srrn's transition behavior is identical to that of a probabilistic finite automaton. 
\end{proof}

\begin{table*}[htb]
\caption{A summary of dataset and experiment characteristics. The values in parentheses are the number of positive sequences.}
\label{tab:dataset}
\small
\begin{tabular}{lllllll}
\toprule
Task         & Architecture & Units & $k$           & Train       & Valid    & Test \\ 
\midrule
Tomita 1     & \srgru       & 100   & 5, 10, 50     & 265 (12)        & 182 (4)      & --        \\
Tomita 2     & \srgru        & 100   & 10, 50        & 257 (6)         & 180 (2)     & --       \\
Tomita 3     & \srgru        & 100   & 50            & 2141 (1028)     & 1344 (615)   & --       \\
Tomita 4     & \srgru        & 100   & 50            & 2571 (1335)     & 2182(1087)   &   --     \\
Tomita 5     & \srgru        & 100   & 50            & 1651 (771)     & 1298(608)   &    --      \\
Tomita 6     & \srgru        & 100   & 50            & 2523 (1221)     & 2222(1098)   &   --       \\
Tomita 7     & \srgru        & 100   & 50            & 1561 (745)     & 680(288)   &    --      \\
BP (large)   & \srlstm(-\textsc{P}) & 100   & 5             & 22286 (13025)   & 6704 (3582)  & 1k     \\
BP (small)   & \srlstm(-\textsc{P}) & 100   & 2,5,10,50,100 & 1008 (601)      & 268 (142)    & 1k     \\
Palindrome   & \srlstm(-\textsc{P}) & 100   & 5             & 229984 (115040) & 50k(25k) & 1k     \\
IMDB (full)  & \srlstm(-\textsc{P}) & 256   & 2,5,10        & 25k           & --            & 25k    \\
IMDB (small) & \srlstm(-\textsc{P}) & 256   & 2,5,10        & 25k           & --            & 25k    \\
MNIST (normal)        & \srlstm(-\textsc{P})   & 256   & 10,50,100     & 60k           & --        & 10k     \\
MNIST (perm.)        & \srlstm(-\textsc{P})    & 256   & 10,50,100     & 60k           & --        & 10k     \\
Fashion-MNIST        & \srlstm(-\textsc{P})    & 256   & 10    & 55k           & 5k        & 10k     \\ 
Copying Memory & \srlstm(-\textsc{P})    & 128, 256   & 5,10,20    & 100k           & --        & 10k     \\ 
Wikipedia& \srlstm(-\textsc{P})    & 300   & 1000    & 22.5M           & 1.2M        & 1.2M     \\  
\bottomrule
\end{tabular}
\end{table*}

\begin{table*}[!htb]
\caption{The lengths ($l$) and depths ($d$) of the sequences in the training, validation, and test sets of the various tasks.}
\label{tab:length}
\begin{tabular}{llll}
\toprule
Task         & Train $l$ \& $d$            & Valid $l$ \& $d$       & Test $l$ \& $d$       \\ 
\midrule
Tomita 1 -7    & $l$ = 0$\sim$13, 16, 19, 22 & $l$ = 1, 4,..., 28     & --                \\
BP (large)   & $d \in [1,5] $        & $d \in [6,10]$   & $d \in [1,20]$   \\
BP (small)   & $d \in [1,5] $        & $d \in [6,10]$   & $d \in [1,20]$   \\
Palindrome   & $l \in [1,25] $       & $l \in [26,50] $ & $l \in [50,500]$ \\
IMDB (full)  & $l \in [11, 2820], l_{aver}=285$                    & --                &$l \in [8, 2956], l_{aver}=278$                \\
IMDB (small) & $l=10$                & --                & $l\in[100,200]$  \\
MNIST(normal)         & $l=784$                 & $l=784$            & $l=784$  \\
MNIST(perm.)         & $l=784$                 & $l=784$            & $l=784$  \\
Fashion-MNIST        & $l=784$                 & $l=784$            & $l=784$          \\ 
Copying Memory  & $l=100,500$                 & --            & $l=100,500$ \\ 
Wikipedia     & $l=22.5M$           & $l=1.2M$        & $l=1.2M$ \\  
\bottomrule
\end{tabular}
\end{table*}

\subsection{Theorem \ref{theorem-dfa-equiv}}

\textbf{For $\tau \rightarrow 0$ the state transition behavior of a memory-less \srrn (using equations~\ref{eqn-transition-1} or \ref{eqn-transition-2}) is identical to that of a deterministic finite automaton.}

\begin{proof}
Let us consider the softmax function with temperature parameter 
$\tau$ $$\alpha_i = \frac{ \exp\left( b_i  / \tau\right) }{ \sum_{i=1}^{k} \exp\left(b_i / \tau\right)}$$ for $1 \leq i \leq k$. \srrns use this softmax function to normalize the scores (from a dot product or Euclidean distance) into a probability distribution. First, we show that for $\tau \rightarrow 0^{+}$, that there is exactly one $M \in \{1, ..., k\}$ such that $\alpha_M = 1$ and $\alpha_i = 0$ for all $i \in \{1, ..., k\}$ with $i\neq M$.
Without loss of generality, we assume that there is a $M \in \{1, ..., k\}$ such that $b_M > b_i$ for all $i \in \{1, ..., k\}, i\neq M$. 
Hence, we can write for $\epsilon_1, ..., \epsilon_k > 0$ as shown in equations (\ref{poorf_1}-\ref{poorf_3}).

Now, for $\tau \rightarrow 0$ we have that $\alpha_M \rightarrow 1$ and for all other $i \neq M$ we have that $\alpha_i \rightarrow 0$. Hence, the probability distribution $\bm{\alpha}$ of the \srrn is always the one-hot encoding of a particular centroid.  

By an argument analog to the one we have made for Theorem \ref{theorem-pdfa}, we can prove that for every state $q\in\mathcal{Q}$ and every input token $a \in \Sigma$, the probability distribution $\bm{\alpha}$ of the \srrn is the same irrespective of the previous input sequences and visited states. Finally, by plugging in the one-hot encoding $\bm{\alpha}$ in both equations~\ref{eqn-transition-1} and \ref{eqn-transition-2}, we can conclude that the transition function of a memory-less \srrn is identical to that of a DFA, because we always chose exactly one new state. 
\end{proof}

\section{Implementation Details}

Unless otherwise indicated we always (a) use single-layer RNNs, (b) learn an embedding for input tokens before feeding it to the RNNs, (c) apply \textsc{RMSprop} with a learning rate of $0.01$ and momentum of $0.9$; (d) do not use dropout or batch normalization of any kind; and (e) use state-regularized RNNs based on equations~3\&6 with a temperature of $\tau=1$ (standard softmax). We implemented \srrns with Theano\citep{theano}~\footnote{\url{http://www.deeplearning.net/software/theano/}}. The hyper-parameter were tuned to make sure the vanilla RNNs achieves the best performance. For \srrns we tuned the weight initialization values for the centroids and found  that sampling uniformly from the interval $[-0.5, 0.5]$ works well across different datasets.
Table \ref{tab:dataset} lists some statistics about the datasets and the experimental set-ups. Table \ref{tab:length} shows the length and  nesting depth (if applicable) for the sequences in the train, validation, and test datasets.

\begin{figure*}[!htb]
\centering
\subfloat{{\includegraphics[width=0.425\textwidth,valign=b]{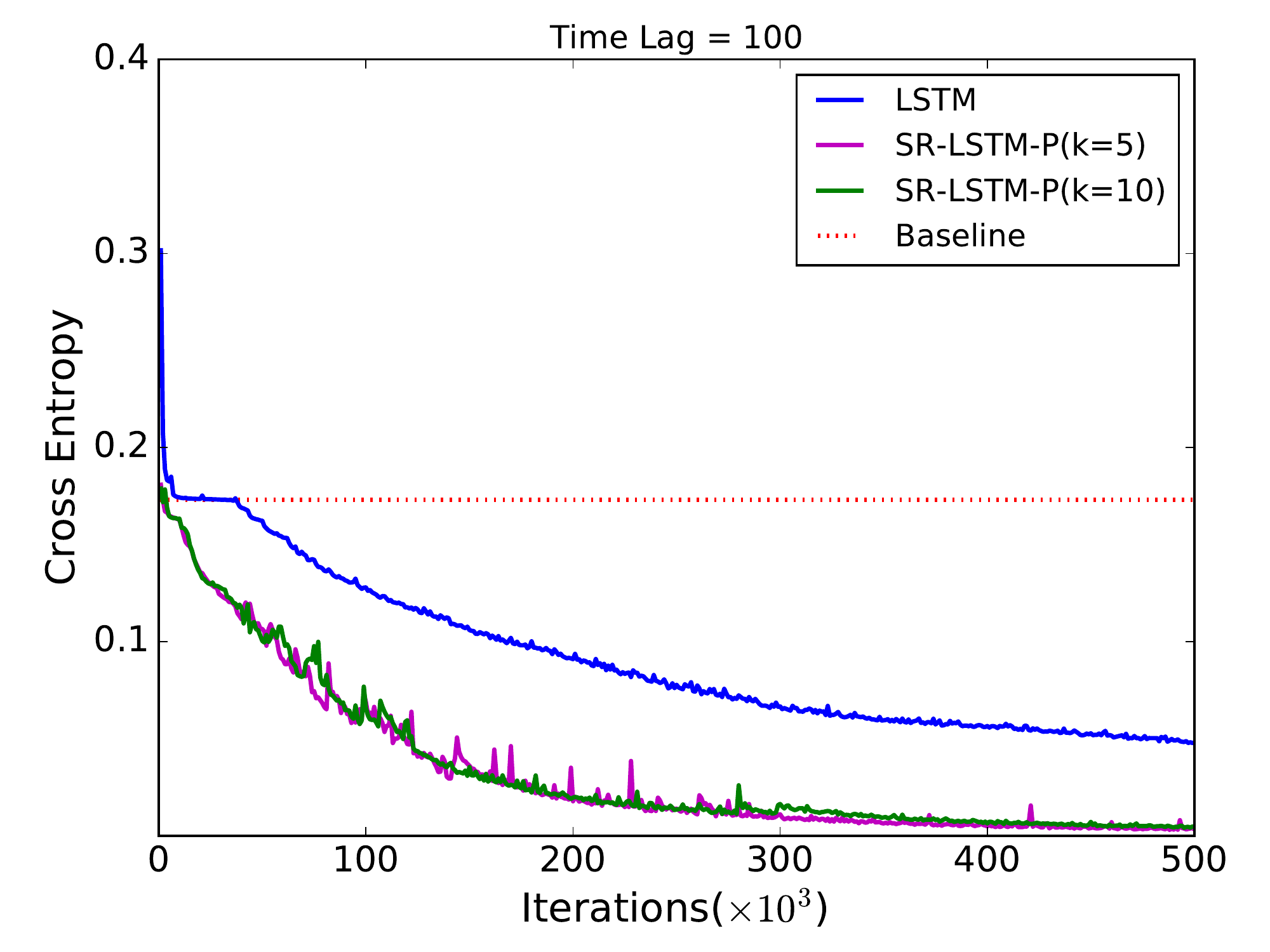} }}%
\hspace{-3mm}
\subfloat{{\includegraphics[width=0.425\textwidth,valign=b]{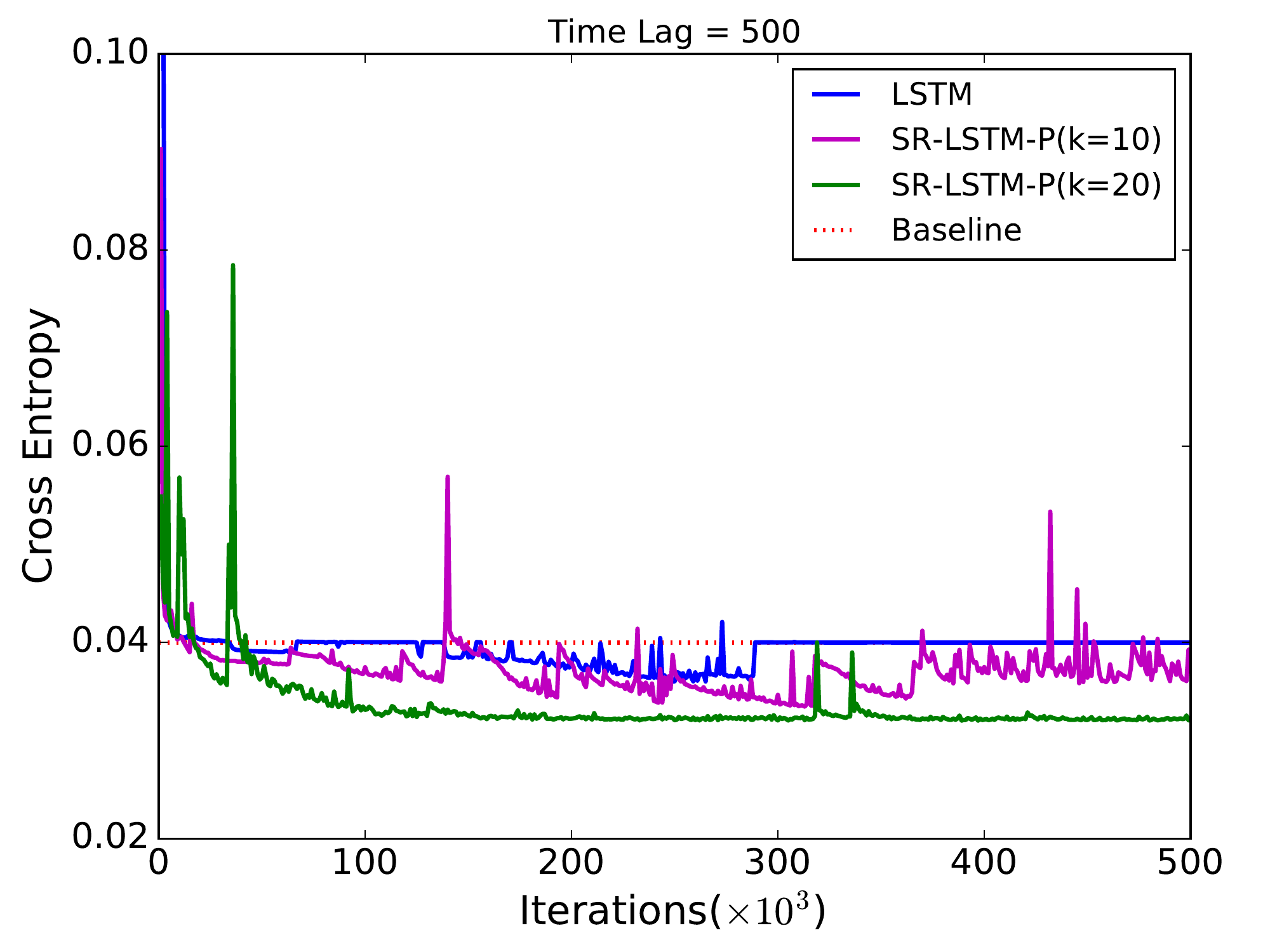} }}%
\caption{\label{fig:cm} The results (test cross entropy loss) on copying memory problem for time lags $T=100$ (left) and $T=500$ (right). We used one recurrent layer with 128 hidden units for $T=100$, and 256 hidden units for $T=500$. }%
\end{figure*}

\section{Experiments on Copying Memory Problem}

Besides the introduced balanced paretness, palindrome task, we also conducted experiment on the copying memory problem~\cite{Hochreiter:1997}. We follow the similar setup as described in~\cite{Hochreiter:1997,arjovsky2016unitary,pmlr-v70-jing17a}.  The alphabet $\Sigma$ has 10 characters $\Sigma =\{c_i\}_{i=0}^9$. The input and output sequences have lengths of $T+2n$, where $T$ is the time lag and $n$ is the length of sequence that need to be memorized and copied. The exemplary input and output are listed as follows.

~~~~~~~~	Input: ~$c_1$$c_4$$c_2$$c_1$$c_8$$c_4$$c_7$$c_3$$c_5$$c_6$\_\_\_\_\_\_\_\_\_$c_9$\_\_\_\_\_ 

~~~~~~~~Output: \_\_\_\_\_\_\_\_\_\_\_\_\_\_$c_1$$c_4$$c_2$$c_1$$c_8$$c_4$$c_7$$c_3$$c_5$$c_6$

We use $n=10$, the first $n$ symbols of input sequence are uniformly sampled from $[c_1,c_8]$, the $c_0$ is used as ``blank'' symbol (as shown as ``\_''  in above examples. The length of ``blank'' sequence is $T$). The $c_9$ is used as an indicator to require algorithms or models to reproduce the first $n$ symbols with exactly same sequential order.
The task is to minimize the average categorical cross entropy at each time step.  Similar to \cite{pmlr-v70-jing17a}, we used 100000 samples for training and 10000 for test. Differently, all RNN models have only one recurrent layer. We used 128 hidden units for $T=100$ and 256 hidden units for $T=500$. The batch size is set to 128. We use \textsc{RMSprop} with a learning rate of $0.001$ and decay of $0.9$. A simple baseline is a \textit{memoryless} strategy, which has categorical cross entropy $\frac{10\log(8)}{T+20}$.

Figure \ref{fig:cm} presents the performance of the proposed method on copying memory problem for $T=100$ and $T=500$. For time log $T=100$, both standard \lstm and \srlstmps are able to beat baseline. Clearly we can see the faster convergence of \srlstmps. For $T=500$, \lstm is hard to beat the baseline, while \srlstmps outperforms baseline in a certain margin.  Both figures demonstrate the memorization capability of \srlstmps over standard \lstm.

\begin{table*}[htb]
\caption{The perplexity results for the \srlstms and  state of the art methods. Here, $\theta_{W+M}$ are the number of model parameters and $\theta_{M}$ the number of model parameters without word representations. $a$ is the attention window size.}
\label{tab:lm}
\begin{tabular}{llllll}
\toprule
Model        & $a$           & $\theta_{W+M}$   & $\theta_{M}$    & Dev   & Test  \\ 
\midrule
RNN          & -        & 47.0M & 23.9M & 121.7 & 125.7 \\
LSTM         &   -       & 47.0M & 23.9M & 83.2  & 85.2  \\ 
FOFE HORNN(3-rd order)\citep{soltani2016higher} &   - & 47.0M & 23.9M & 116.7 & 120.5 \\
Gated HORNN(3-rd order)\citep{soltani2016higher}&  -   & 47.0M & 23.9M & 93.9  & 97.1  \\ 
RM(+tM-g) \citep{tran2016recurrent}            &  15    & 93.7M & 70.6M & 78.2  & 80.1  \\ 
Attention \citep{daniluk2017frustratingly}        &  10        & 47.0M & 23.9M & 80.6  & 82.0  \\
Key-Value  \citep{daniluk2017frustratingly}    &    10         & 47.0M & 23.9M & 77.1  & 78.2  \\
Key-Value Predict  \citep{daniluk2017frustratingly}  & 5      & 47.0M & 23.9M & 74.2  & 75.8  \\
4-gram RNN   \citep{daniluk2017frustratingly}       &    -    & 47.0M & 23.9M & 74.8  & 75.9  \\ 
\midrule
\lstmp             &   -   & 47.0M & 23.9M & 85.8 & 86.9 \\ 
\srlstm($k=1000$)        &    -       & 47.3M & 24.2M & 80.9 & 82.7 \\ 
\srlstmp($k=1000$)       &    -        & 47.3M & 24.2M & 80.2 & 81.3 \\ 
\bottomrule
\end{tabular}
\end{table*}
\section{Experiments on Language Modeling}

We evaluated the  \srlstms on language modeling task with the Wikipedia dataset~\citep{daniluk2017frustratingly}\footnote{The wikipedia corpus is available at https://goo.gl/s8cyYa}. It consists of 7500 English Wikipedia articles. We used the same experimental setup as in previous work~\citep{daniluk2017frustratingly}. We used the provided training, validation, and test dataset:  22.5M words in the training set, 1.2M in the validation, and 1.2M words in the test set. We used the 77k most frequent words from the training set as vocabulary. We report the results in Table \ref{tab:lm}. We use the model with the best perplexity on the on the validation set. Note that we only tuned the number of centroids for \srlstm and \srlstmp and used the same hyperparameters that were used for the vanilla LSTMs.

The results show that the perplexity results for the \srlstm and \srlstmp outperform those of the vanilla LSTM and LSTM with peephole connection. The difference, however, is modest and we conjecture that the ability to model long-range dependencies is not that important for this type of language modeling tasks. This is an observation that has also been made by previous work~\cite{daniluk2017frustratingly}. The perplexity of the \srlstms cannot reach that of state of the art methods. The methods, however, all utilize a mechanism (such as attention) that allows the next-word decision to be based on a number of past hidden states. The \srlstms, in contrast, makes the next-word decision only based on the current hidden state.

\section{Tomita Grammars and DFA Extraction}

The Tomita grammars are a collection of 7 regular languages over the alphabet \{0,1\}~\citep{tomita:cogsci82}. Table \ref{tab:tomita} lists the regular grammars defining the Tomita grammars. 

\begin{table*}
\begin{tabular}{ll}
\toprule
Grammars& Descriptions \\ 
\midrule
 1& 1$^\ast$ \\
 2& (10)$^\ast$ \\
 3&  An odd number of consecutive 1s is followed by an even number of consecutive 0s\\
 4&  Strings not contain a substring ``000''\\
 5&  The numbers of 1s and 0s are even \\
 6&  The difference of the numbers of 1s and 0s is a multiple of 3\\
 7&  0$^\ast$1$^\ast$0$^\ast$1$^\ast$\\ 
 \bottomrule
\end{tabular}
\caption{The seven Tomita grammars~\citep{tomita:cogsci82}.}
\label{tab:tomita}
\end{table*}

We follow previous work \citep{Wang:2007:nc, Schellhammer:1998} to construct the transition function of the DFA (deterministic finite automata). 

\begin{algorithm2e*}[htb]
\algrule[1.15pt]
  \small
  \caption{Computes DFA transition function}
  \label{alg:DFA extraction}
  \KwIn{pre-trained \srrn, dataset $\textbf{D}$, alphabet $\Sigma$, start token $\mathtt{s}$}
  \KwOut{transition function $\delta$ of the DFA}
 % \hrulefill \\
  %\text {$\{c_k\}_{k=1}^K \leftarrow \srrn$} \hfill \# obtain centroids of the \srrn \\
  \For{$i,j \in \{1, ..., k\} \mbox{ and all } x \in \Sigma$}   
  {
      \text{$\bm{\mathcal{T}}[(c_{i}, x_t, c_j)] = 0$} \hfill \# initialize transition counts to zero \\
  }
  \text{$\{p_i\}_{i=1}^k \leftarrow \srrn(\mathtt{s})$} ~\hfill~\# compute the transition probabilities for the start token \\
  \text{$j = \argmax_{i \in \{1, ..., k\}} (p_i)$}  ~\hfill~\# determine $j$ the centroid with max transition probability \\
  \text{$c_0 = j $} ~\hfill~\# set the start centroid $c_0$ to $j$  \\
 %\text \# for each sample in testset \\
  \For{$\bfx = (x_1, x_2,..., x_T) \in \textbf{D}$}   
  {
   %\text \# compute the transition probs \\
     
     \For{$t \in [1, ..., T]$}   
     {
         \text{$\{p_j\}_{j=1}^k \leftarrow \srrn(x_{t})$} ~\hfill~\# compute the transition probabilities for the $t$-th token \\
         \text{$j = \argmax_{i \in \{1, ..., k\}} (p_i)$}  ~\hfill~\# determine $j$ the centroid with max transition probability \\
         \text{$c_t = j $} ~\hfill~\# set $c_t$, the centroid in time step $t$, to $j$  \\
         \text{$\bm{\mathcal{T}}[(c_{t-1}, x_t, c_t)] \leftarrow \bm{\mathcal{T}}[(c_{t-1}, x_t, c_t)] + 1$ }   \hfill \# increment transition count\\   
     }
  }
\For{$i \in \{1, ..., k\} \mbox{ and } x \in \Sigma$}
{
    \text{$\delta(i, x) = \argmax_{j \in \{1, ..., k\}}  \bm{\mathcal{T}}[(i, x, j)]$} ~\hfill~\# compute the transition function of the DFA \\
}
\Return{$\delta$} \\
\algrule[1.15pt]
\end{algorithm2e*}

%\vspace{-2mm}

We follow earlier work \citep{pmlr-v80-weiss18a} and attempt to train a GRU to reach 100\% accuracy for both training and validation data.  We first trained a single-layer \gru with $100$ units on the data. We use GRUs since they are memory-less. Whenever the \gru converged within 1 hour to a training accuracy of $100\%$, we also trained a \srgru based on equations~3\&6 with $k=50$ and $\tau=1$.  This was the case for the grammars 1-4 and 7. For grammar 5 and 6, our experiments show that both vanilla GRU and \srgru were not able to achieve 100\% accuracy. In this case, \srgru (97.2\% train and 96.8\% valid accuracy) could not extract the correct DFA for grammar 5 and 6.  The exploration of deeper GRUs and their corresponding \srgrus (2 layers as in \citep{pmlr-v80-weiss18a} for DFA extraction could be interesting future work.  

\begin{figure}[htb]
\centering
\includegraphics[width=0.8\textwidth]{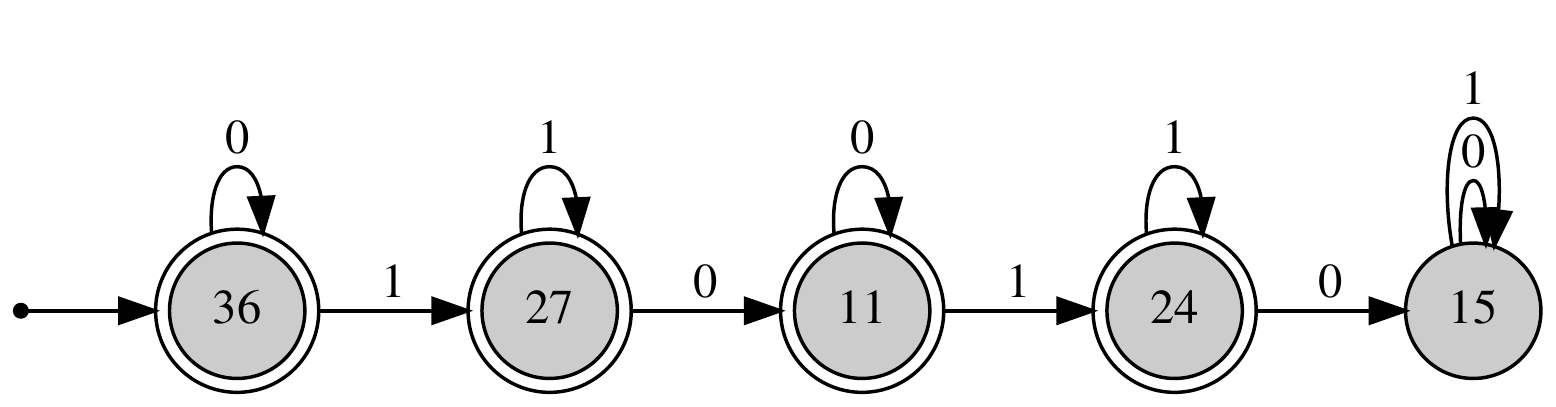}%
\caption{The DFA extracted from the \srgru for Tomita grammar 7. The numbers inside the circles correspond to the centroid indices of the \srgru. Double circles indicate accept states.}%
\label{fig:grammar_7}
\end{figure}

Algorithm \ref{alg:DFA extraction} lists the pseudo-code of the algorithm that constructs the transition function of the DFA. Figure \ref {fig:grammar_7} shows the extracted DFA for grammar 7. All DFA visualization in the paper are created with GraphViz \footnote{\url{https://www.graphviz.org/}}. 

\begin{figure*}[!htb]
\vspace{-5mm}
\centering
\subfloat{{\includegraphics[width=0.375\textwidth,valign=b]{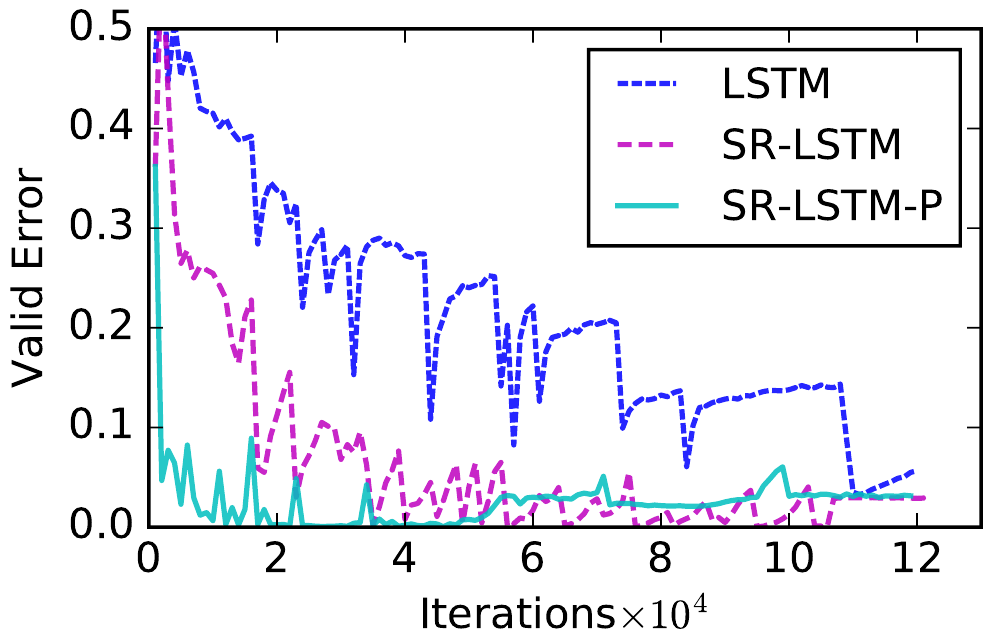} }}%
\hspace{1mm}
\subfloat{{\includegraphics[width=0.375\textwidth,valign=b]{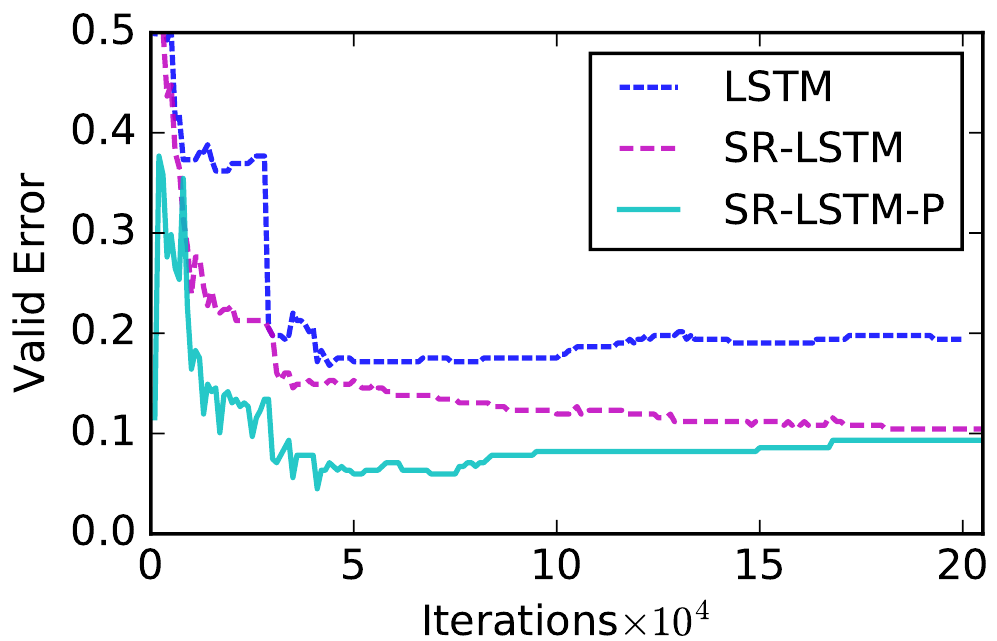} }}%
\caption{\label{fig:bp_learning} The error curves on the validation data for the \lstm, \srlstm, and \srlstmp ($k=5$)on the large BP dataset (left) and the small BP dataset (right). }%
\end{figure*}
\begin{figure*}[!htb]
\vspace{-5mm}
\centering
\subfloat{{\includegraphics[width=0.365\textwidth,valign=b]{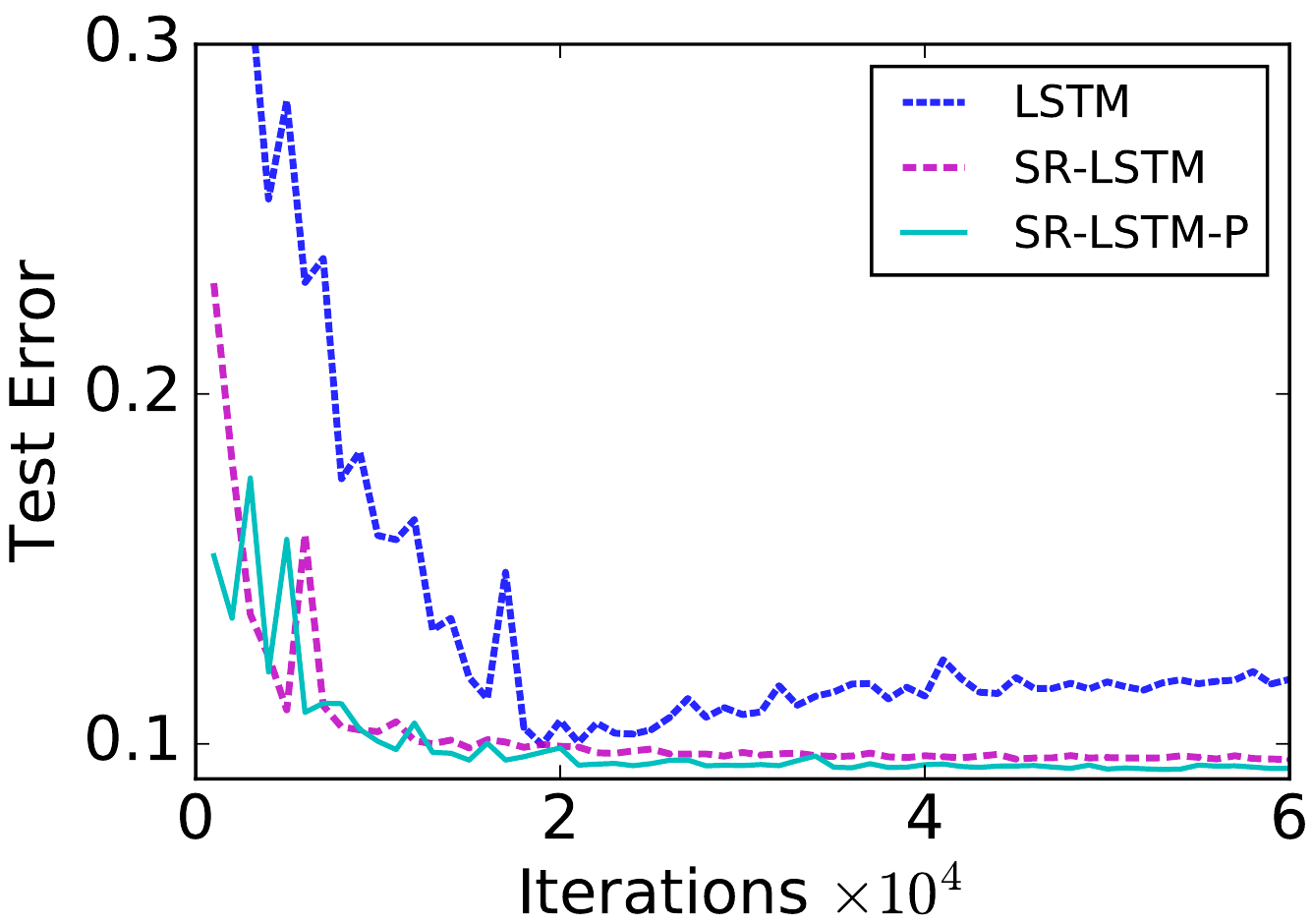} }}%
\hspace{1mm}
\subfloat{{\includegraphics[width=0.365\textwidth,valign=b]{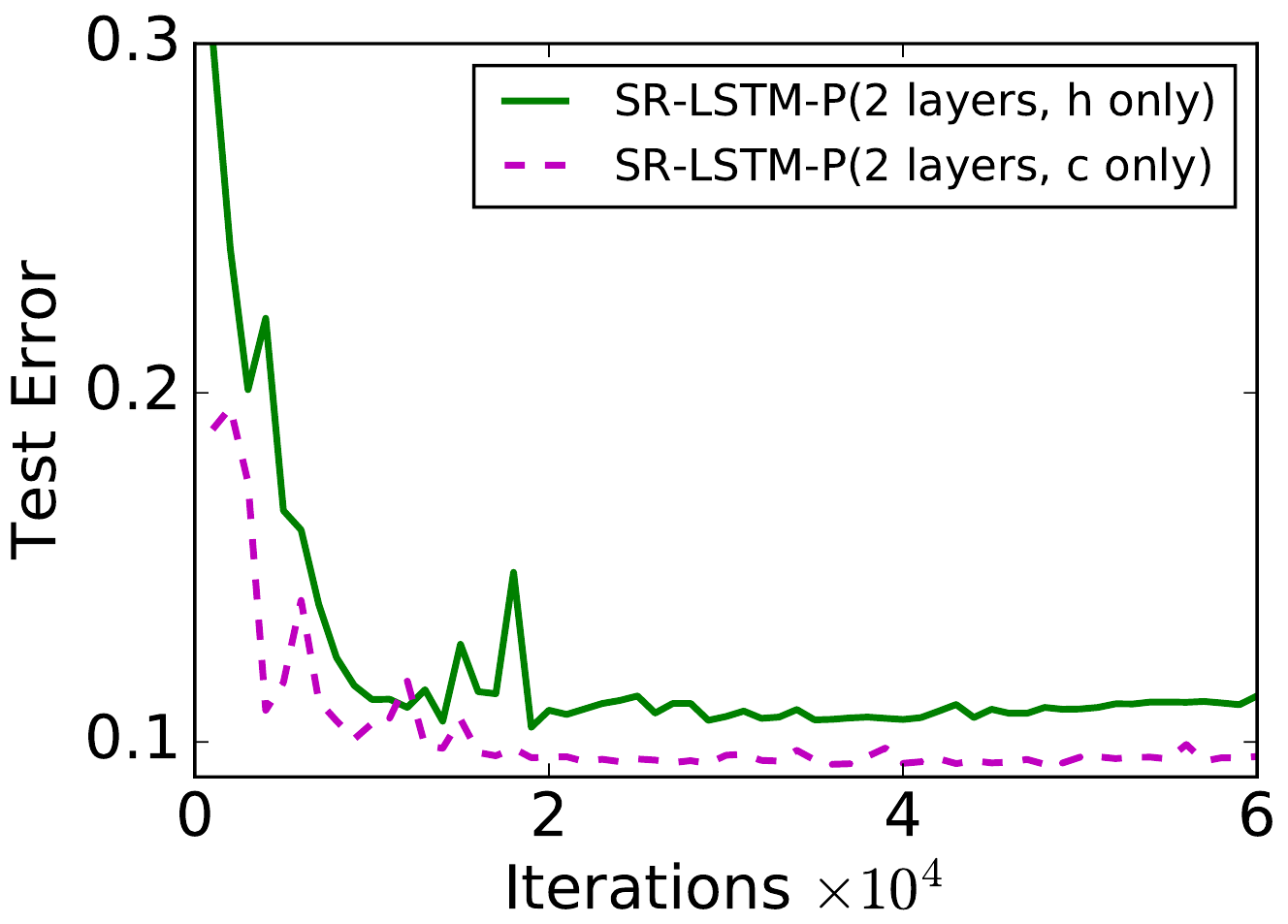} }}%
\caption{\label{fig:imdb_learning} The error curves on the test data for the \lstm, \srlstm, \srlstmp ($k=10$) on the IMDB sentiment analysis dataset. (Left) It shows state-regularized RNNs show better generalization ability. (Right) A 2-layer \srlstmp achieves better error rates when the classification function only looks at the last cell state compared to it only looking at the last hidden state. }%
\end{figure*}
\begin{figure*}[!htb]
\subfloat[Vector $\bfh_t$ of the \lstm]{{\includegraphics[width=0.45\textwidth,valign=b]{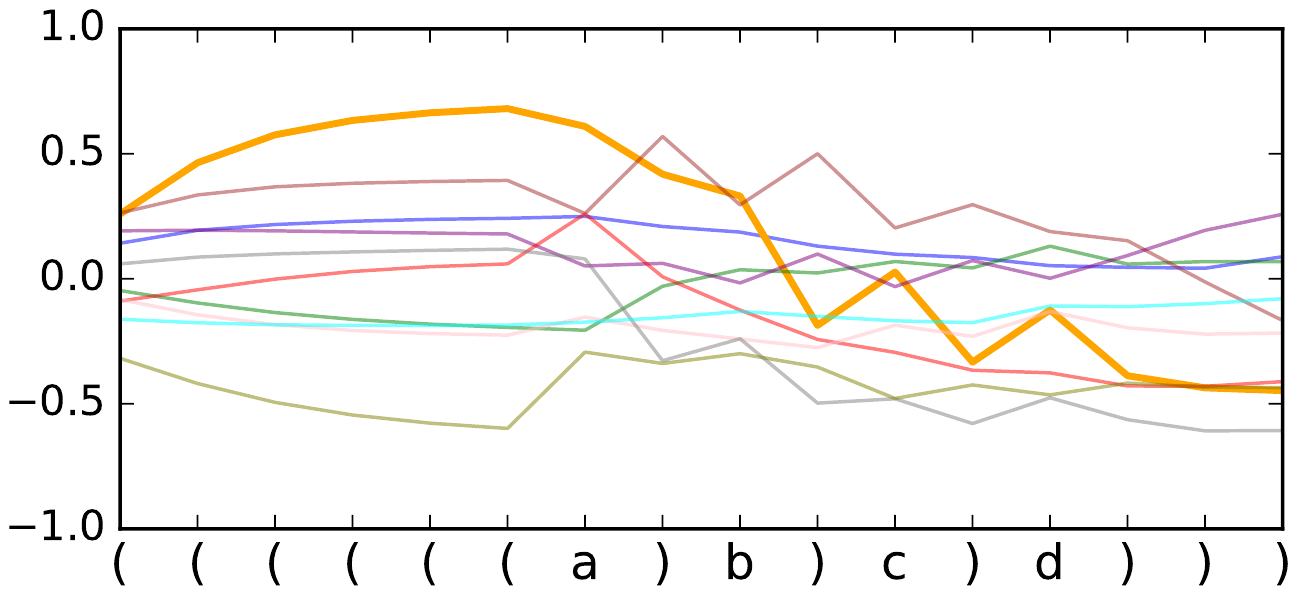} } \hspace{5mm} }% 
\subfloat[Vector $\bfc_t$ of the \lstm]{{\includegraphics[width=0.44\textwidth,valign=b]{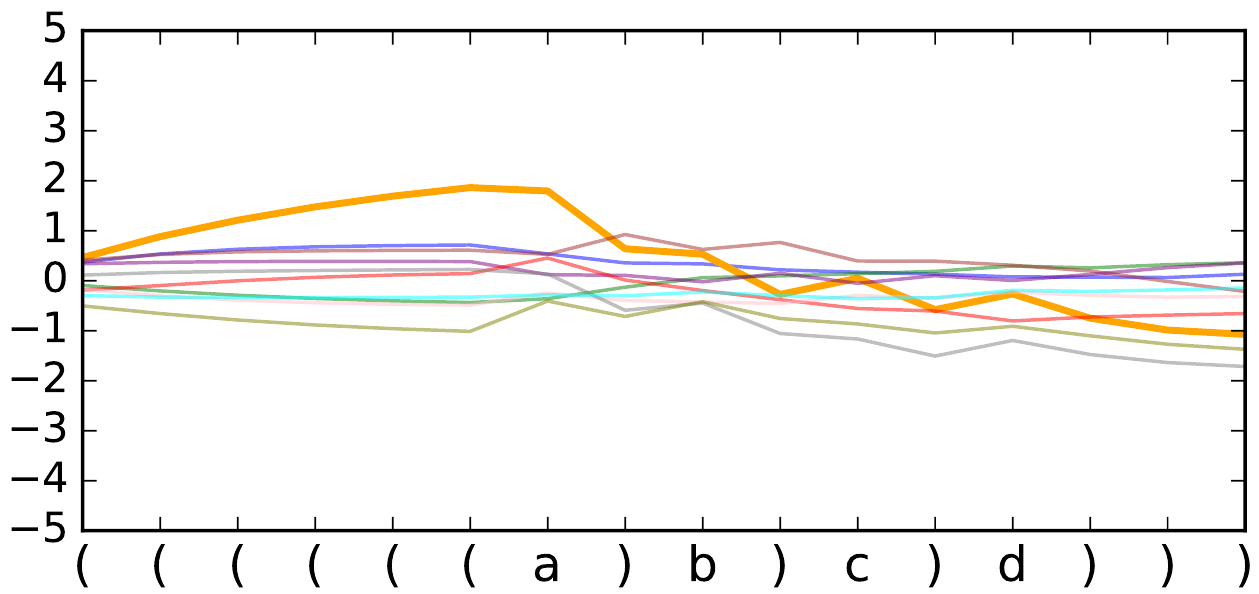} }} \\
\subfloat[Vector $\bfh_t$ of the \srlstmp]{{\includegraphics[width=0.45\textwidth,valign=b]{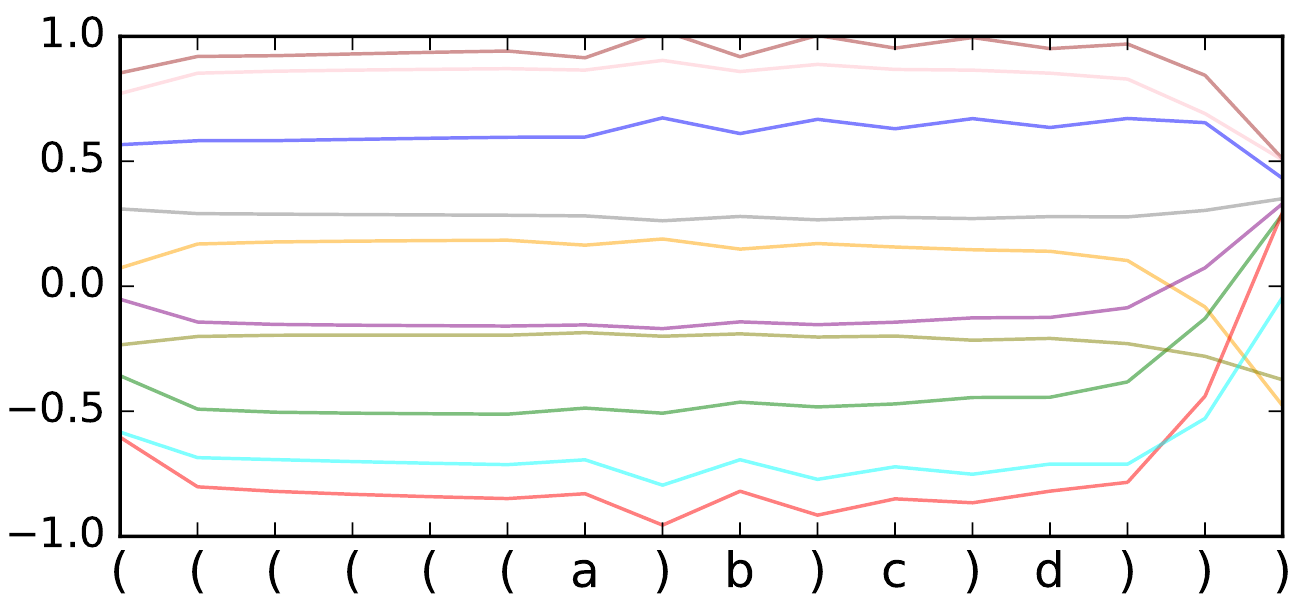} } \hspace{5mm}  }% 
\subfloat[Vector $\bfc_t$ of the \srlstmp]{{\includegraphics[width=0.44\textwidth,valign=b]{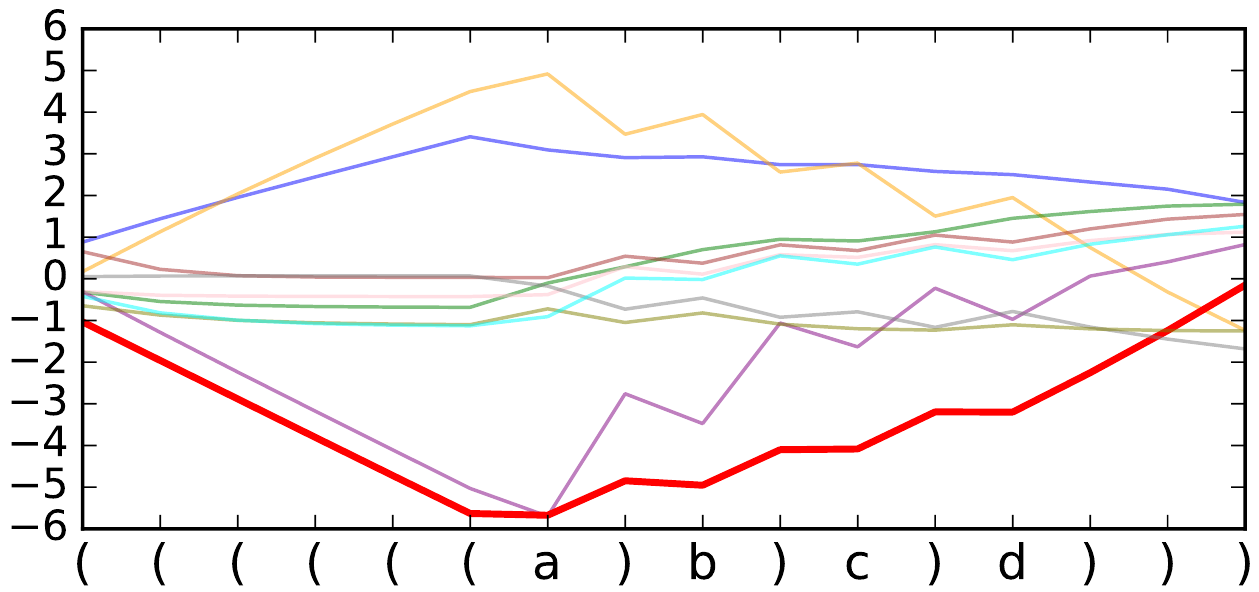} }}% 
\caption{\label{fig:bp_vis_2} Visualization of hidden state $\bfh_t$ and cell state $\bfc_t$ of the \lstm and the \srlstmp for a specific input sequence from BP.  Each color corresponds to one of 10 hidden units. The \lstm memorizes the number of open parentheses both in the hidden and to a lesser extent in the cell state (bold yellow lines). The memorization is not accomplished with saturated gate outputs and a drift is observable for both vectors. The \srlstmp maintains two distinct hidden states (accept and reject) and does not visibly memorize counts through its hidden states. The cell state is used to cleanly memorize the number of open parentheses (bold red line) with saturated gate outputs ($\pm1$). A state vector drift is not observable (the solutions with less drift to generalize better~\cite{gers2001lstm}).}%
\end{figure*}
\section{Training Curves}

Figure \ref{fig:bp_learning} plots the validation error during training of the \lstm, \srlstm, and \srlstmp on the BP (balanced parentheses) datasets. Here, the \srlstm and \srlstm both have $k=5$ centroids. The state-regularized LSTMs tend to reach better error rates in a shorter amount of iteration. 

Figure \ref{fig:imdb_learning} (left) plots the test error of the \lstm, \srlstm, and \srlstmp on the IMDB dataset for sentiment analysis. Here, the \srlstm and \srlstm both have $k=10$ centroids. In contrast to the LSTM, both the  \srlstm and the \srlstm do not overfit. 

Figure \ref{fig:imdb_learning} (right) plots the test error of the \srlstmp when using either (a) the last hidden state and (b) the cell state as input to the classification function. As expected, the cell state contains also valuable information for the classification decision. In fact, for the \srlstmp it contains more information about whether an input sequence should be classified as positive or negative.

\section{Visualization, Interpretation, and Explanation}

The stochastic component and its modeling of transition probabilities and the availability of the centroids facilitates novel ways of visualizing and understanding the working of \srrns. 

\begin{figure*}[!htb]
\centering
\includegraphics[width=0.85\textwidth]{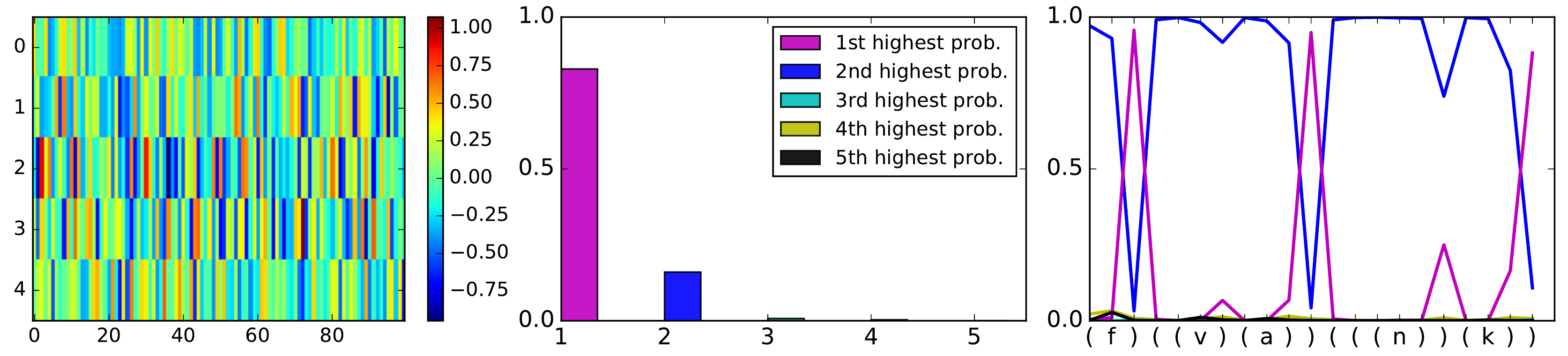}%
\caption{Visualization of a \srlstmp with $k=5$ centroids, a hidden state dimension of $100$, trained on the large BP data. (Left) visualization of the learned centroids. (Center)  mean transition probabilities when ranked highest to lowest.  This shows that the transition probabilities are quite spiky. (Right) transition probabilities for a specific input sequence.}%
\label{fig:bp_centroids}
\end{figure*}
\begin{figure*}[htb]
\centering
\includegraphics[width=0.6\textwidth,valign=b]{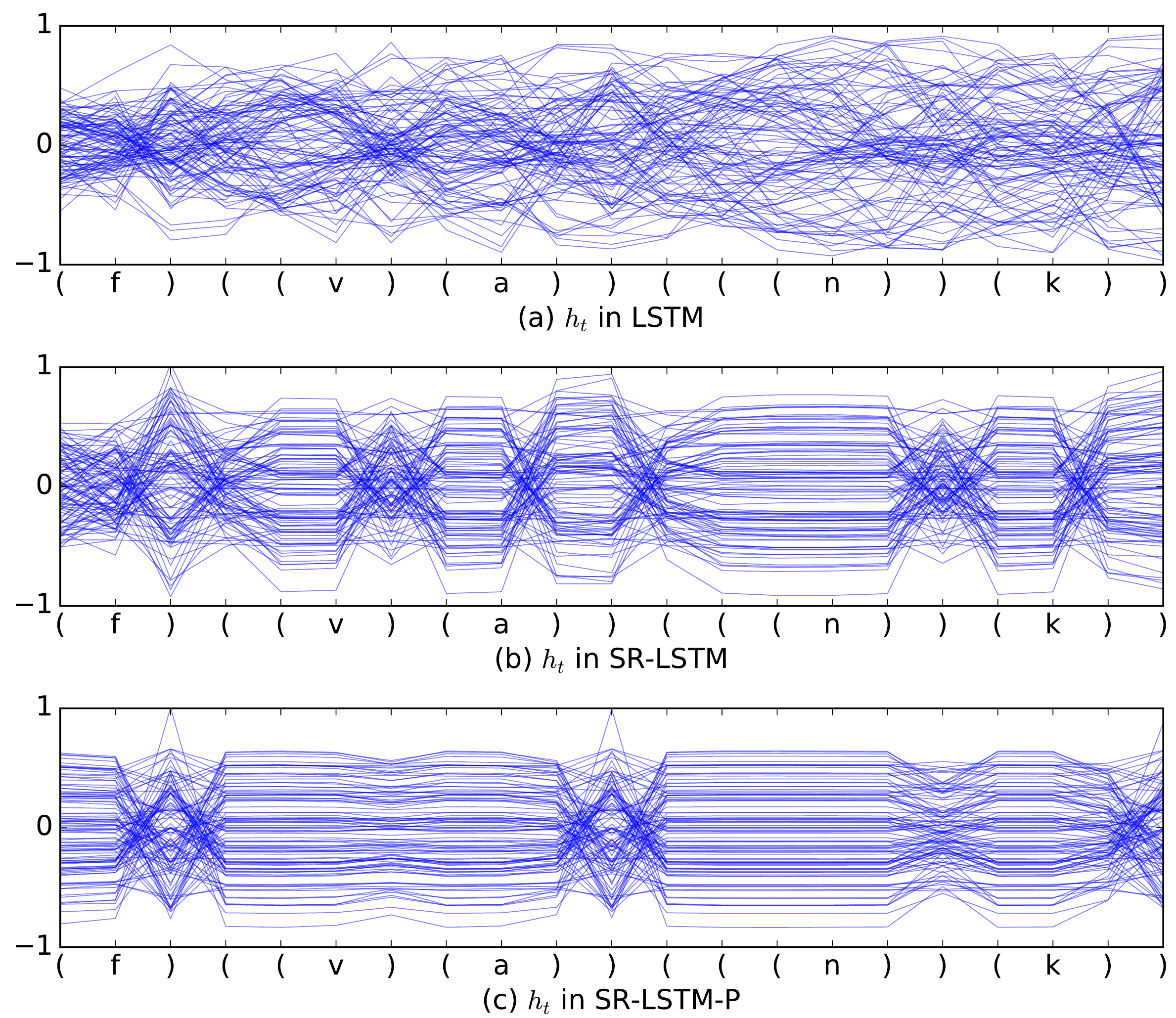}%
\caption{Visualizations of the hidden states $\bfh$ of a vanilla \lstm , an \srlstm, and a \srlstmp ($k=10$) for a specific input sequence. All models were trained on BP and have hidden state dimension of $100$. The LSTM memorizes with its hidden state. The \srlstm and \srlstmp utilize few states and have a more stable behavior over time. }%
\label{fig:bp_h_c_100_1}
\end{figure*} 

\begin{table*}[!htb]
\begin{tabular}{lr}
\hline
Centroids                         & Top-5 words with probabilities                                                                                        \\ \hline \hline
centroid 0                        & piece (0.465)    instead (0.453)    slow (0.453)    surface (0.443)    artificial (0.37)           \\
centroid 1                        & told (0.752)    mr. (0.647)    their (0.616)    she (0.584)    though (0.561)                      \\
centroid 2                        & just (0.943)    absolutely (0.781)    extremely (0.708)    general (0.663)    sitting (0.587)      \\
{\color[HTML]{FD6864} centroid 3} & {\color[HTML]{FD6864} worst (1.0)    bad (1.0)    pointless (1.0)    boring (1.0)    poorly (1.0)} \\
centroid 4                        & jean (0.449)    bug (0.406)    mind (0.399)    start (0.398)    league (0.386)                     \\
centroid 5                        & not (0.997)    never (0.995)    might (0.982)    at (0.965)    had (0.962)                         \\
centroid 6                        & against (0.402)    david (0.376)    to (0.376)    saying (0.357)    wave (0.349)                   \\
centroid 7                        & simply (0.961)    totally (0.805)    c (0.703)    once (0.656)    simon (0.634)                    \\
{\color[HTML]{036400} centroid 8} & {\color[HTML]{036400} 10 (0.994)    best (0.992)    loved (0.99)    8 (0.987)    highly (0.987)}   \\
centroid 9                        & you (0.799)    strong (0.735)    magnificent (0.726)    30 (0.714)    honest (0.69)                \\ \hline
\end{tabular}
\caption{\label{tab:imdb_vis_10} List of prototypical words for the $k=10$ centroids of an \srlstmp trained on the IMDB dataset. The top-5 highest transition probability words are listed for each centroid. We colored the positive centroid words in green and the negative centroid words in red.}
\label{tab:imdb_10}
\end{table*}
\subsection{Balanced Parentheses}
Figure \ref{fig:bp_vis_2} presents the visualization of hidden state $\bfh_t$ and cell state $\bfc_t$ of the \lstm and the \srlstmp for a specific input sequence from BP.

Figure \ref{fig:bp_centroids} (left) shows the $k=5$ learned centroids of a \srlstmp with hidden state dimension $100$. 

Figure \ref{fig:bp_centroids} (center) depicts the average of the ranked transition probabilities for a large number of input sequences. This shows that, on average, the transition probabilities are spiky, with the highest transition probability being on average $0.83$, the second highest $0.16$ and so on. 

Figure \ref{fig:bp_centroids} (right) plots the transition probabilities for a \srlstmp with $k=5$ states and hidden state dimension $100$ for a specific input sequence of BP.

Figure \ref{fig:bp_h_c_100_1} visualizes the hidden states $\bfh$ of a LSTM, \srlstm, and \srlstmp trained on the large BP dataset. The \srlstm and \srlstmp have $k=5$ centroids and a hidden state dimension of $100$. One can see that the LSTM memorizes with its hidden states. The evolution of its hidden states is highly irregular. The \srlstm and \srlstmp, on the other hand, have a much more regular behavior. The \srlstmp utilizes mainly two states to accept and reject an input sequence.

\begin{figure*}[!htb]
\centering
\includegraphics[width=\textwidth,valign=b]{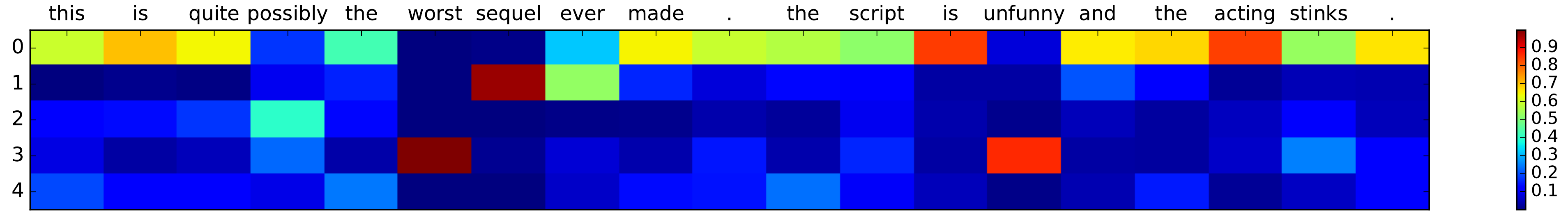}%
\caption{Visualization of the transition probabilities for an \srlstmp with $k=5$ centroids trained on the IMDB dataset for a negative input sentence.}%
\label{fig:imdb_probs}
\end{figure*} 

\begin{figure*}[!htb]
\centering
\includegraphics[width=\textwidth,valign=b]{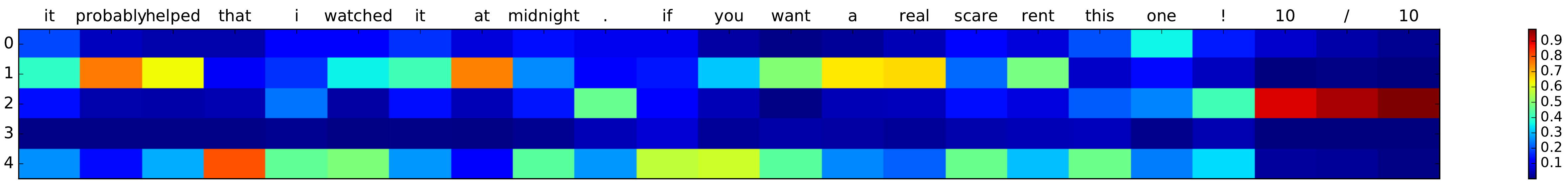}%
\caption{Visualization of the transition probabilities for an \srlstmp with $k=5$ centroids trained on the IMDB dataset for a positive input sentence.}%
\label{fig:imdb_probs_1}
\end{figure*} 

\subsection{Sentiment Analysis}

Since we can compute transition probabilities in each time step of an input sequence, we can use these probabilities to generate and visualize prototypical input tokens and the way that they are associated with certain centroids. We show in the following that it is possible to associate input tokens (here: words of reviews) to centroids by using their transition probabilities. 

For the sentiment analysis data (IMDB) we can associate words to centroids for which they have high transition probabilities. To test this, we fed all test samples to a trained \srlstmp. We determine the average transition probabilities for each word and centroid and select those words with the highest average transition probability to a centroid as the prototypical words of said centroid. Table \ref{tab:imdb_10} lists the top 5 words according to the transition probabilities to each of the 5 centroids for the \srlstmp with $k=10$. It is now possible to inspect the words of each of the centroids to understand more about the working of the \srrn. 

Figure \ref{fig:imdb_probs} and \ref{fig:imdb_probs_1} demonstrate that it is possible to visualize the transition probabilities for each input sequence. Here, we can see the transition probabilities for one positive and one negative sentence for a \srlstmp with $k=5$ centroids.

In addition, \srrn can be used to visualize the transition probabilities for specific inputs. To explore this,  we trained an \srlstmp ($k=10$) on the MNIST (accuracy 98\%) and Fashion MNIST (86\%) data \citep{xiao2017fashion}, having the models process the images pixel-by-pixel as a large sequence. 
%Figure \ref{fig:mnist_explain} visualizes the transition probabilities with a heatmap for specific input images.

\begin{figure}[!htb]
\centering
\subfloat{{\includegraphics[width=0.95\textwidth,valign=b]{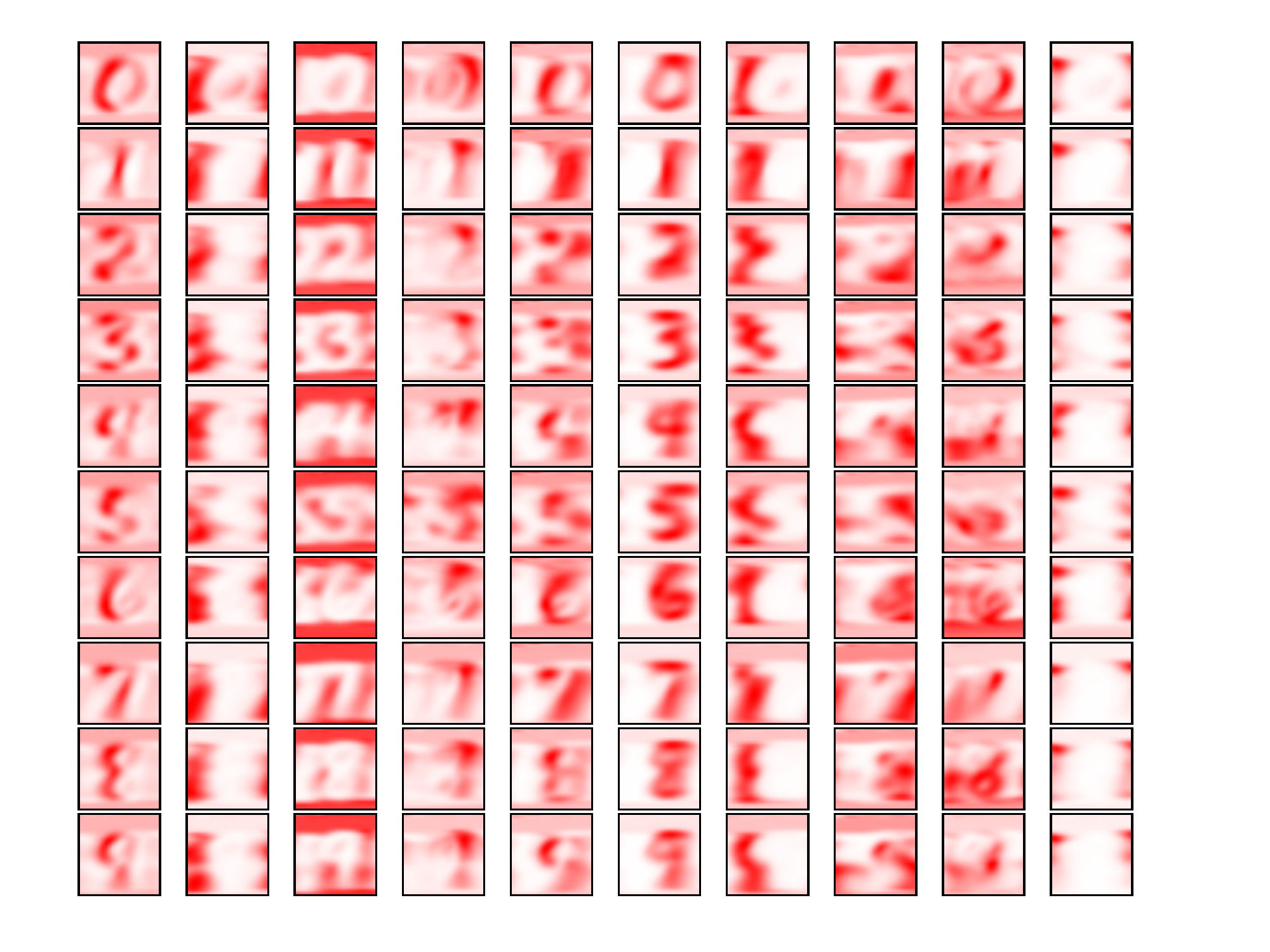} }}%
\caption{\label{fig:minst_proto} Visualization of average transition probabilities of the \srlstmp with $k=10$ centroids, over all test images. Each row represents a digit class (a concept) and each column depicts the prototype (average transition probability) for each of the centroids.}%
\end{figure}
\begin{figure*}[!pt]
\centering
\subfloat{{\includegraphics[width=0.45\textwidth,valign=b]{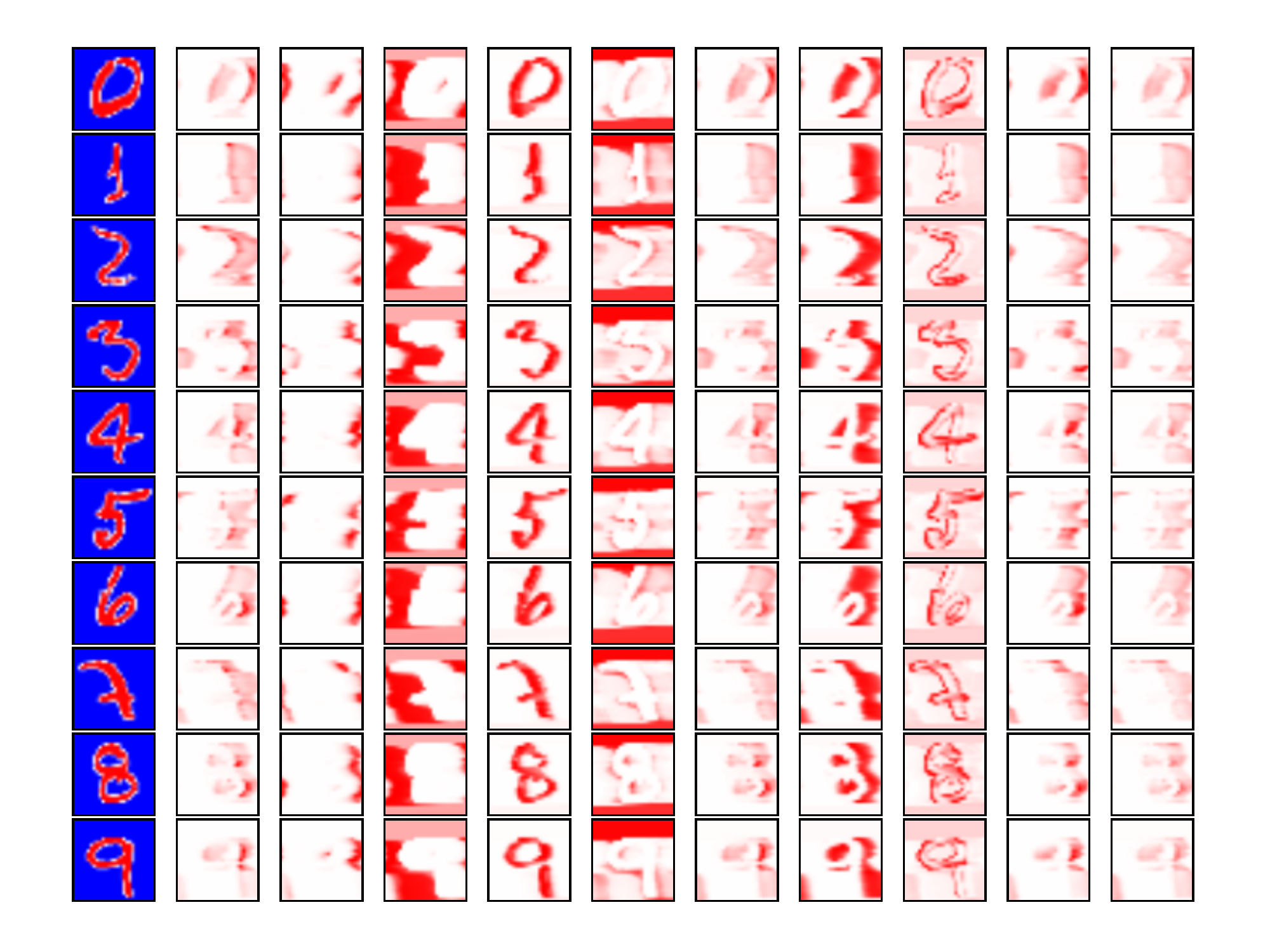} }}%
\hspace{1mm}
\subfloat{{\includegraphics[width=0.45\textwidth,valign=b]{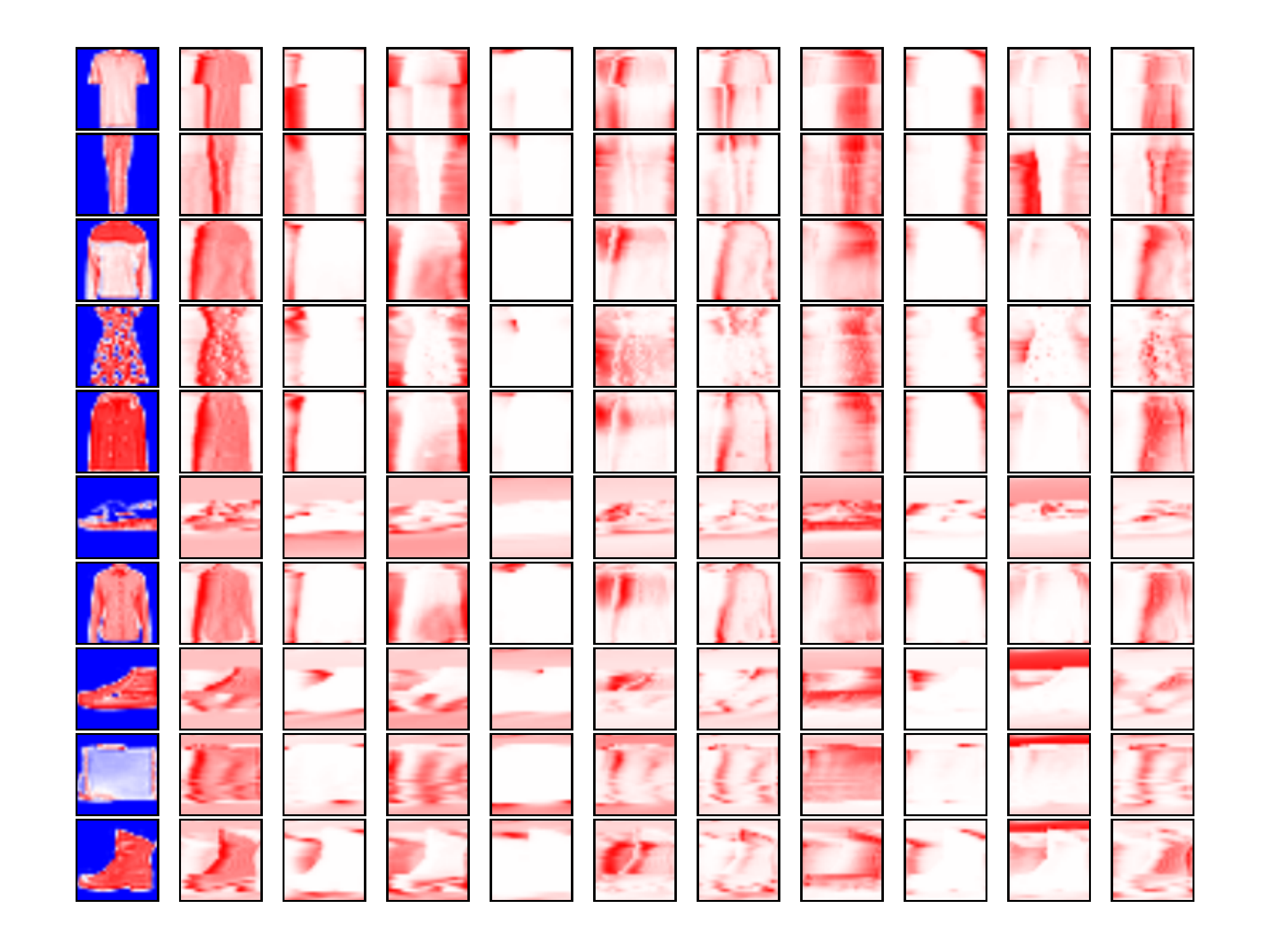} }}%
\caption{\label{fig:mnist_explain} We can visualize the working of an \srlstmp on a specific input image by visualizing the transition probabilities of each of the centroids (here: $k=10$). (Top) The visualizetion for some MNIST images. (Bottom) The visualization for some Fashion-MNIST images. The first column depicts the input image and the second to to $11^{th}$ the state transition probability heatmaps corresponding to the 10 centroids.}%
\end{figure*}
\subsection{MNIST and Fashion-MNIST}

For pixel-by-pixel sequences, we can use the \srrns to directly generate prototypes that might assist in understanding the way \srrns work. We can compute and visualize the average transition probabilities for all examples of a given class. Note that this is different to previous post-hoc methods (e.g., activation maximization \citep{berkes2006analysis,nguyen2016synthesizing}), in which a  network is trained first and in a second step a second neural network is trained to generate the prototypes.  

Figure \ref{fig:minst_proto} visualizes the prototypes (average transition probabilities for all examples from a digit class) of \srlstmp for $k=10$ centroids. One can see that each centroid is paying attention to a different part of the image. 

Figure \ref{fig:mnist_explain} visualizes the explanations that generated for RNN predictions with a given input of \srlstmp for $k=10$ centroids. One can see that the important features are highlighted for explaining the predictions.

\end{appendices}

\end{document}